\documentclass[11pt]{article}
\pdfoutput=1

\usepackage{arxiv}

\usepackage[utf8]{inputenc} % allow utf-8 input
\usepackage[T1]{fontenc}    % use 8-bit T1 fonts
\usepackage{hyperref}       % hyperlinks
\usepackage{url}            % simple URL typesetting
\usepackage{booktabs}       % professional-quality tables
\usepackage{amsfonts}       % blackboard math symbols
\usepackage{nicefrac}       % compact symbols for 1/2, etc.
\usepackage{microtype}      % microtypography
\usepackage{lipsum}

\usepackage{microtype}
\usepackage{graphicx}
\usepackage{subfigure}
\usepackage{booktabs} % for professional tables
\usepackage{amsmath} % for equation
\usepackage{algorithm}
\usepackage[noend]{algpseudocode}

\title{Off-Policy Adversarial Inverse Reinforcement Learning}

\author{
  Samin Yeasar Arnob \\
  Department of Electrical and Computer Engineering\\
  McGill University\\
  \texttt{samin.arnob@mail.mcgill.ca} \\
}

\begin{document}
\maketitle

\begin{abstract}
Adversarial Imitation Learning (AIL) is a class of algorithms in Reinforcement learning (RL), which tries to imitate an expert without taking any reward from the environment and does not provide expert behavior directly to the policy training. Rather, an agent learns a policy distribution that minimizes the difference from expert behavior in adversarial setting. Adversarial Inverse Reinforcement Learning (AIRL) leverages the idea of AIL, integrates a reward function approximation along with learning the policy and shows the utility of IRL in the transfer learning setting. But the reward function approximator that enables transfer learning does not perform well in imitation tasks. We propose an Off-Policy Adversarial Inverse Reinforcement Learning (Off-policy-AIRL) algorithm which is sample efficient as well as gives good imitation performance compared to the state-of-the-art AIL algorithm in the continuous control tasks. For the same reward function approximator, we show the utility of learning our algorithm over AIL by using the learned reward function to retrain the policy over a task under significant variation where expert demonstrations are absent.

\end{abstract}

% keywords can be removed
\keywords{Reinforcement Learning \and Inverse Reinforcement Learning \and Transfer Learning}

% =========================
%========================================================
\section{Introduction}
%=======================================================
\label{introduction}

Reinforcement learning (RL) is a very useful framework for learning complex behavior in control, where an agent interacts with an environment and tries to learn the expected behavior by optimizing an objective function that ensures the highest cumulative reward over specific time steps. A reward is often human-engineered function that tries to ensure higher rewards for expected behavior within the environment. Designing a reward function for a complex environment can be tricky. Poorly designed reward function may lead to a fatal mistake when applied in real-world applications. 

Generative Adversarial Imitation Learning (GAIL) \cite{GAILHoE16} introduces AIL where a generator (policy) is used to compute trajectory and a binary classifier called discriminator classifies generated behavior from the expert. As imitation learning algorithms only learn policy, they miserably fail when there is a change in environmental dynamics.

Inverse reinforcement learning (IRL) \cite{AndrewNgDisentangledreward,AndrewFeatureMatching} addresses the importance of learning reward function and thus learn reward function approximator along with policy. Adversarial Inverse Reinforcement Learning (AIRL) \cite{AIRLfu2018} proposes an IRL algorithm in adversarial learning, which shows a promising result when there is considerable variability in the environment from the demonstration setting. But compared to AIL algorithms, AIRL does not perform well in imitation performance \cite{DACkostrikov2018,AIRLfu2018}. It is also important to note the best performing reward function approximator configuration for imitation and transfer learning tasks in AIRL algorithm are different. In our work we perform transfer learning using the same configuration that works best in imitation performance.

The paper makes the following contributions: (i) we show the scope of using a stochastic policy (SAC) in DAC algorithm, and it gives improved imitation performance (ablation study added on the appendix Figure \ref{DAC_TD3vsSAC}). (ii) Ee propose an Off-Policy Adversarial Inverse Reinforcement Learning algorithm (off-policy-AIRL) which is sample efficient and improves imitation performance of prior IRL algorithms through comparing imitation performance with state-of-the-art AIL algorithm on continuous control tasks. (iii) We show utility of learning IRL algorithm over imitation by reusing the learned reward function to train policy under certain changes in dynamics where expert demonstrations are absent. We use \emph{Soft Actor-Critic} (SAC), which makes our algorithm more sample efficient. (iv) We also show using Multiplicative Compositional Policies (MCP) allows more flexibility in retraining policy in transfer task and at the same time improves imitation performance.

%==========================

%=========================
\section{Background}
\label{background}

We define a Markov decision process (MDP) as a tuple $(S,A,T,r,\gamma,\rho_0)$, where $S$,$A$ are the state and action space respectively, $\gamma \in [0,1)$ is the discount factor. We consider the dynamic or transition distribution $T(s'|s,a)$, initial state distribution $\rho_0(s)$ and reward function $r(s,a)$ are unknown in IRL setup and can be learned through interaction with the MDP.

AIL evolved from maximum casual entropy IRL framework \cite{MaximumCasualIntropyziebart2010}, which considers an entropy regularized MDP with the goal to find the optimal policy $\pi*$ that maximizes the expected entropy-regularized discounted reward under $\pi$,$T$ and $\rho_0$:

\begin{equation}
    \pi^* = \arg \max_{\pi} \mathbf{E}_{\tau \sim \pi} \Big[ \sum_{t=0}^T \gamma^t r(s_t,a_t) + H (\pi(.|s_t)) \Big].
\end{equation}

\noindent Here $\tau = (s_0,a_0...s_T,a_T)$ defines the sequence of states and actions induced by policy and  $H (\pi(.|s_t))$ is the discounted causal entropy of policy $\pi$.  In imitation task, expert demonstrations $D = {\tau_1, \tau_2 ... \tau_n}$ are given and an agent is expected to produce trajectory which are similar to the expert and infer reward function. Under certain constrain of feature matching \cite{MaximumEntropyIRL} derived following objective function: 
\begin{equation}
\begin{split}
P_\psi(\tau) & \propto e^{R(\tau)} = \frac{1}{z} e^{R(\tau)}, \label{maximum_en} \\
            & \propto p(s_0) \prod p(s_{t+1}|s_t,a_t) e^{\gamma^t r_\psi(s_t,a_t)},
\end{split}
\end{equation}
where $z = \int_\tau e^{R(\tau)}$ is the partition function and can be expressed as probability of trajectories with equal reward/cost are equal likely and probability of trajectory with higher rewards are exponentially more likely. Equation \eqref{maximum_en} can be interpreted as solving a maximum likelihood problem: $max_{\psi} \mathbf{E}_{\tau \sim D}[ \log p_\psi (\tau)]$. But the assumptions that are made in \cite{MaximumCasualIntropyziebart2010} are model / transition function $T(s'|s,a)$ is known and applied on small state-action space. 
% == GAIL

GAIL \cite{GAILHoE16}, a model-free algorithm, is the first framework to draw connection between generative adversarial network (GAN) and imitation learning which works in continuous state-action space. GAIL uses a generative model or generator, $G$ that acts as the policy. Purpose of the generator is to learn policy $\pi_\theta$ such that the state-action visitation frequency $\rho_{\pi_\theta}$ is similar to expert's $\rho_{\pi_E}$ without directly being provided with the expert demonstrations. A discriminator $D$ is a binary classifier which tries to distinguish the data distribution of the generator $\rho_{\pi}$ from the expert's $\rho_{\pi_E}$. Objective for a generator is to minimize the difference with expert visitation frequency, while a discriminator wants to maximize the difference and thus derived the entropy regularized objective as following:
\begin{equation}
    distance_{\min_{\pi_\theta} \max_{D_\psi}} (\rho_\pi, \rho_{\pi_E}) 
    = \nonumber  \min_{\pi_\theta} \max_{D_\psi}  \Big[ \mathbf{E}_{\pi_\theta} [\log D_\psi(s,a)]  + \nonumber  \mathbf{E}_{\pi_E} [\log(1-D_\psi(s,a))] -\lambda H(\pi_\theta) \Big], 
\end{equation}
\noindent where entropy term $H$ is policy regularizer controlled by $\lambda \geq 0$. GAIL uses two separate neural-networks to represent generator and discriminator. For any given state generator tries to take expert like action. Discriminator takes state-action pairs as input and computes the likelihood of the input coming from an expert. Generator uses a reward function $ R(s,a) = -\log D_\psi(s,a)$ in order to train it's network parameters. 

GAIL is sample efficient in terms of number of expert demonstrations required but not quite sample efficient in terms of required interaction with the environment to learn the policy.
Thus GAIL is not suitable for many real-world applications. \emph{Discriminator Actor critic} (DAC) \cite{DACkostrikov2018} uses Twin-delayed deep deterministic actor-critic (TD3) \cite{TD3}, an off-policy algorithm as generator to improve upon the sample efficiency of existing methods, and it extends the learning environment with absorbing states. DAC criticises AIL algorithms (i.e GAIL) for inducing reward biases by using strict positive or negative reward function and proposes a reward function $ r(s,a) = \log (D_\theta(s,a)) - \log (1-D_\theta(s,a))$ in order to achieve unbiased rewards. Combined, these changes remove the survival bias and solves sample inefficiency problem of GAIL.

% =================== break ============================

% === IRL
For being an imitation learning algorithm, instead of learning cost/reward function, both GAIL and DAC only recover expert policy and thus will fail miserably in dynamic environments. A discriminator function, that learns policy as well as the cost/reward function, is proposed by \cite{GAN-GCLFinnCAL16} by optimizing following objective for discriminator network: 
\begin{equation}
    D_\psi = \frac{p_\psi(\tau)}{p_\psi(\tau)+q(\tau)} \overset{(a)}{=} \frac{1/z*e^{R_\psi}}{1/z*e^{R_\psi} + q(\tau)}.
\end{equation}

And thus,
\begin{equation}
    \mathcal{L}_{dis} = \max_{D_\psi} \Big[ E_{\tau \sim p}[ \log D_\psi(\tau)] + \nonumber E_{\tau \sim q}[ \log (1-D_\psi(\tau))] \Big],
\end{equation}

\noindent where $p(\tau)$ is the data distribution and $q(\tau)$ is the policy distribution and (a) consider derived objective function \eqref{maximum_en} from \cite{MaximumCasualIntropyziebart2010}. But \cite{GAN-GCLFinnCAL16} contains no experimental demonstration of imitation performance. Later on AIRL \cite{AIRLfu2018} shows, when we consider a whole trajectory to compute discriminator update it suffers from high variance issue and that leads to a poor performing generator. Instead AIRL computes discriminator update using $(s,a,s')$ tuple: 

\begin{equation}
\begin{split}
    D_{\psi,\omega} & = \frac{ p_{\psi,\omega}(s,a,s') }{p_{\psi,\omega}(s,a,s') + q(s,a)}, \\
                    & = \frac{ e^{f_{\psi,\omega}(s,a,s')} }{ e^{f_{\psi,\omega}(s,a,s')} + q(s,a)}, \\
                     & = \frac{ e^{f_{\psi,\omega}(s,a,s')} }{ e^{f_{\psi,\omega}(s,a,s')} + \pi_\theta(a|s)},
\end{split}
\end{equation}

\noindent where $f_{\psi,\omega}(s,a,s') = r_\psi (s,a) + \gamma \Phi_\omega(s') - \Phi_\omega(s)$ is a potential-based reward function, $r_\psi (s,a)$ is a reward function approximator and $ \Phi_\omega$ is reward shaping term controlled by the dynamics. In \cite{AIRLfu2018} $f_{\psi,\omega}(s,a,s')$ is referred as disentangled reward function \cite{Ng:1999:PIU:645528.657613} due to $r_\psi(s)$ being indifferent to the changes in dynamics $T$.

%=========================

%============================
% =================================== %

\section{Off-policy Adversarial Inverse Reinforcement Learning}
\label{off_policy_airl}

Off-policy-AIRL algorithm is inspired from DAC \cite{DACkostrikov2018}, which is an adversarial imitation learning algorithm. DAC addresses two existing problem with AIL algorithms, which are reward function bias and sample inefficiency. Reward functions used in these algorithms induce an implicit bias by using strict positive or negative reward functions, which may work for some environments but in many cases become a reason for sub-optimal behavior.  

Furthermore, \cite{DACkostrikov2018} shows adversarial methods improperly handle terminal states. This introduces implicit reward priors that can affect the policy`s performance. In particular, many imitation learning implementations \cite{GMMIL,GAILHoE16,AIRLfu2018} and MDPs omit absorbing states $s_a$. Thus they implicitly assign 0 reward to terminal/absorbing states and adds a bias to the reward learning process. Thus we use the same approach to learn the reward for absorbing states. Return of the terminal state is defined as $R_T=r(s_T,a_T)+\sum_{t=T+1}^{\infty} \gamma^{t-T}r(s_a,.)$, where $r(s_a,0)$ is learned reward, instead of just $R_T=r(s_T,a_T)$. To make the implementation more stable, the terminal reward is analytically derived using the following equation:
\begin{equation}
    R_T=r(s_T,a_T)+\gamma \frac{r(s_a,.)}{1-\gamma} 
\end{equation}

Like any other imitation learning algorithm DAC only learns the policy. We want to use the best of both worlds and thus replace the binary discriminator of DAC and use formulated discriminator function from AIRL \cite{AIRLfu2018} to learn a reward function approximator along with the policy. From \cite{AIRLfu2018} it is evident that AIRL provides a poor imitation performance compared to GAIL. It is important to note that the best performing policy in \cite{AIRLfu2018} for imitation task is trained using state-action dependent reward function $r_{\psi}(s,a)$, which fails to re-train policy in the transfer learning task. For state dependent reward function $r_{\psi}(s)$ \cite{AIRLfu2018} successfully completes transfer learning task but it comes with a cost of poor imitation performance. In this work, we perform transfer learning using the same configuration that works best in imitation performance. We improve upon the prior IRL algorithm performance and while still being sample efficient by using Soft-Actor-Critic (SAC) \cite{SAC} as the generator.

% ===================================================
We store our experience in a replay buffer $\mathcal{R}$ to utilize them during the off-policy update of the generator. After each time the environment reaches a terminal condition and resets, we update our discriminator and generator for the same episodic timesteps. We use an expert buffer $\mathcal{R_E}$ to store the expert trajectory and sample the $(s_E,a_E,s_E{'})$ pair in batch while update our discriminator.

\subsection{Discriminator update rule}
% ======================================

During discriminator update we randomly sample mini-batch of state-action-next-state pairs ${(s,a,.,s')}_b$ from both expert buffer $\mathcal{R_E}$ and replay buffer $\mathcal{R}$. We use following equation derived in \cite{AIRLfu2018} to compute the discriminator output,
\begin{equation}
    D_{\psi,\omega} = \frac{ e^{f_{\psi,\omega}(s,a,s')} }{ e^{f_{\psi,\omega}(s,a,s')} + \pi_\theta(a|s)}.
\label{airl_disc_eq}
\end{equation}

\noindent where,
\begin{equation}
\begin{split}
    f_{\psi,\omega} & = r_\psi(s) + \gamma \Phi_\omega(s') - \Phi_\omega(s), \\
                     & =  r_\psi(s) + \gamma V_\omega(s') - V_\omega(s).
\label{disentanged_reward_eq}
\end{split}
\end{equation}

\noindent Here reward approximator $r_\psi$ can be a function of state, state-action or state-action-next-state and $\Phi$ can be any function that gives a measure of being at any state $s$. Similar to \cite{AIRLfu2018} we use value function $V_\omega$ as reward shaping terms. We find the best imitation performance using state dependent reward function. 

Output of the discriminator will predict likelihood of being an expert and thus objective of the discriminator is to minimize following binary cross entropy loss:
\begin{equation}
    min \ \mathcal{L}_{\psi,\omega} =  \ min \ \sum_{b=1}^B [- \log D_{\psi,\omega}(s_b,a_b,s_b^{'}) - \log (1-D_{\psi,\omega}(s_{b_E},a_{b_E},s_{b_E}^{'} ))].
\end{equation}

With in few iterations of the discriminator update, it easily can classify the expert from the generator data. Thus to make the generator learning more stable we use gradient penalty \cite{gp, DACkostrikov2018}.

% =================================
\subsection{Generator update rule}
% =================================
%  
In IRL setting, we consider the actual reward from the environment is non-existent and rather formulate a reward approximator to train policy. AIRL \cite{AIRLfu2018} algorithm uses a on-policy policy-gradient algorithm as its generator and proposes following reward approximator $\hat{r}(s,a,s')$ to update the policy:

\begin{equation}
    \hat{r}_{\psi,\omega}(s,a,s') = \log D_{\psi,\omega} (s,a,s') - \log (1-D_{\psi,\omega}(s,a,s')). \label{disentanged_reward_eq2} 
\end{equation}

Using equation \eqref{airl_disc_eq} it is also referred as entropy regularized reward function:

\begin{equation}
\begin{split}
    \hat{r}_{\psi,\omega}(s,a,s') & = \log \frac{e^{f_{\psi,\omega}(s,a,s')} }{e^{f_{\psi,\omega}(s,a,s')}+ \pi(a|s)} - \log \frac{\pi_\theta(a|s)}{e^{f_{\psi,\omega}(s,a,s')}+ \pi_\theta(a|s)}, \\ \nonumber
                            & =  f_{\psi,\omega}(s,a,s') - \log \pi_\theta(a|s).  
    \label{disentanged_reward_eq3}
\end{split}
\end{equation}

For improving imitation performance and sample efficiency we use SAC as our generator. SAC being an off-policy algorithm, samples $(s,a,s',done)$ tuple from buffer when trains the network.
For our algorithm we experiment different reward approximator which is discussed in following section and use trained reward approximator to update critic or $Q_\phi$ function in SAC with entropy term through following gradient step:
\begin{equation}
\nabla_\phi J_Q(\phi) =  \nabla_\theta Q_\phi(s_t,a_t)\Big[ Q_\phi(s_t,a_t)- \hat{r}_{\psi,\omega}(s,a,s') + \nonumber \gamma Q_{\phi'} (s_{t+1},a_{t+1})- \alpha \log \pi_\theta(a_{t+1}|s_{t+1}) \Big].
\end{equation}

\noindent Similar to \cite{TD3,SAC} we update actor or policy $\pi_\theta$ for every second update of our $Q_\phi$ function using following gradient step:
\begin{equation}
\nabla_\theta [E_{a_t \sim \pi_\theta} [\alpha \log \pi_\theta(a_t|s_t) -Q_\phi(s_t,a_t) ]].
\end{equation}

% =========================

\subsection{Reward function selection}

Implementation of AIRL discriminator \cite{AIRLfu2018} does not work off the shelf as now we have to train an off-policy generator. We experiment on different way to compute the reward function for the generator update.

In AIRL, the author uses the value function of the states $V_\omega(s)$ to measure additional shaping terms. We experiment on different variations of reward approximator $\hat{r}_{\psi,\omega}$ to update policy and disentangled reward function $f_{\psi,\omega}$ to compute discriminator output so that we find the combination of these two approximators that gives the best performance in off-policy-AIRL.

% I have used policy and discriminator update rules same as DAC while optimizing the objective of AIRL. 
%It doesn't use any squashing function at the output of the discriminator $D_{\theta,\phi}$. But in my implementation this makes the learning unstable. To resolve the issue I have bounded the output of discriminator with \emph{Sigmoid} squashing function for all following experiments and it seems to solve the instability issue

\textbf{Implementation 1}:
%==============================
\begin{itemize}
    \item Policy update using :  $\hat{r}_{\psi,\omega}(s,a,s') = \log D_{\psi,\omega} (s,a,s') -  \log (1-D_{\psi,\omega}(s,a,s'))$
    \item Disentangled reward function :  $f_{\psi,\omega} = r_\psi(s,a) + \gamma V_\omega(s') - V_\omega(s)$
\end{itemize}

\textbf{Implementation 2}
% =============================
\begin{itemize}
    \item Policy update using :  $\hat{r}_{\psi,\omega}(s,a,s')= \log D_{\psi,\omega} (s,a,s') -  \log (1-D_{\psi,\omega}(s,a,s')) $
    \item Disentangled reward function :  $f_{\psi,\omega} = r_\psi(s) + \gamma V_\omega(s') - V_\omega(s)$
\end{itemize}

\textbf{Implementation 3}
% =============================
\begin{itemize}
    \item Policy update using : $ \hat{r}_{\psi,\omega}(s,a,s') = r_{\psi}(s,a)$ 
    \item Disentangled reward function : $f_{\psi,\omega} = r_\psi(s,a) + \gamma V_\omega(s') - V_\omega(s)$
\end{itemize}

\textbf{Implementation 4}
% ============================
\begin{itemize}
    \item Policy update using : 
    $\hat{r}_{\psi,\omega}(s,a,s')= r_{\psi}(s)$ 
    \item Disentangled reward function : 
    $f_{\psi,\omega} = r_\psi(s) + \gamma V_\omega(s') - V_\omega(s)$
\end{itemize}

\noindent For implementation (1) and (2), we use the exact policy update rule which is used in AIRL \cite{AIRLfu2018}. In implementation (1) we consider a state dependent reward function $r_\psi(s)$ and for implementation (2) we consider reward to be a function of state-action pair $r_\psi(s,a)$. As we see from equation \eqref{disentanged_reward_eq2}, reward computed to update policy is equivalent to entropy regularized reward function. We evaluate the disentangled reward function for both state dependent $r_\psi(s)$ and state-action dependent $r_\psi(s,a)$. For implementation (3) and (4) we follow similar comparative study where policy update discard entropy regularization and use direct output from our reward approximator $r_\psi$. As discussed in DAC \cite{DACkostrikov2018}, strict negative or positive reward approximator induces bias. Similar to DAC, our reward approximator $r_\psi$ gives both positive and negative rewards thus does not suffer reward biasing.

%============================

% ================ Transfer learning ========================

\section{Transfer learning task}

Advantage of IRL over imitation learning is that we can leverage the learned reward function to retrain a new policy. If the reward function captures the underlying objective of an agent, then it is possible to use the reward function under robust changes of the dynamics. Imitation learning successfully learn policy in the training domain but it is not possible to use the policy when there is a significant domain shift \cite{AIRLfu2018}. We use the learned reward function from off-policy-AIRL algorithm to re-train our policy in transfer learning setting. To the best of our knowledge AIRL \cite{AIRLfu2018} is the only IRL framework that shows utility of reward function in transfer learning for continuous control tasks and thus we demonstrate our experiments using same environments from \cite{AIRLfu2018} (see Figure \ref{transfer_task_envs}) and we also consider expert demonstrations to be absent for the transfer tasks.

For transfer learning experiments we consider dynamic changes in following two criterion, where either (1) dynamics of the agent or (2) dynamics of the environment changes while the goal/objective of the agent and action dimension remain the same. Through theses experiments we show the learned reward function $r_{\psi}$ indeed can help the policy to behave under dynamic changes even when expert behavior is completely absent. For transfer learning experiments we use the same hyper-parameter and update rules described in prior.

\subsection{Criterion:1}

By changing dynamics of an agent, we refer to a significant structural change of an agent. To evaluate transfer learning performance under this criterion we use \emph{Customized Ant} \cite{AIRLfu2018} environment. In both imitation and transfer task the quadrupedal Ant requires to run forward, while dynamics of the agent during test session is changed by disabling and shrinking two legs. In MuJoCo setting, \cite{AIRLfu2018} sets the gear ratios to 1, whereas the other joints were set to 150. The gear ratio is a parameter which translates actions to torques on the joints, so the disabled joints are $\sim150x$ more difficult to move. This significantly changes its gait. In a MDP this puts a restriction in action space $A$, which results in a new transition dynamics $T'(s|s,a)$. Thus optimal policy $\pi^*$ is changed even though the objective and action-dimension remains the same.

\subsection{Criterion:2}

By changing dynamics of the environment, we consider notable variation in the trained environment such that it directly results in a different transition dynamics $T'(s|s,a)$ without putting restriction in action-space $A$. Objective of \emph{Shifting Maze} \cite{AIRLfu2018} task is to reach to a specific goal position, where during imitation learning, agent is trained to go around the wall on the left side, but during transfer task environment setup is changed and must go around the wall on the right. Thus optimal policy $\pi^*$ is changed even though the objective and the dynamics of the agent remains same.

% ====== figures

\begin{figure}[hbt!]
\centering
    \includegraphics[width=0.2\linewidth]{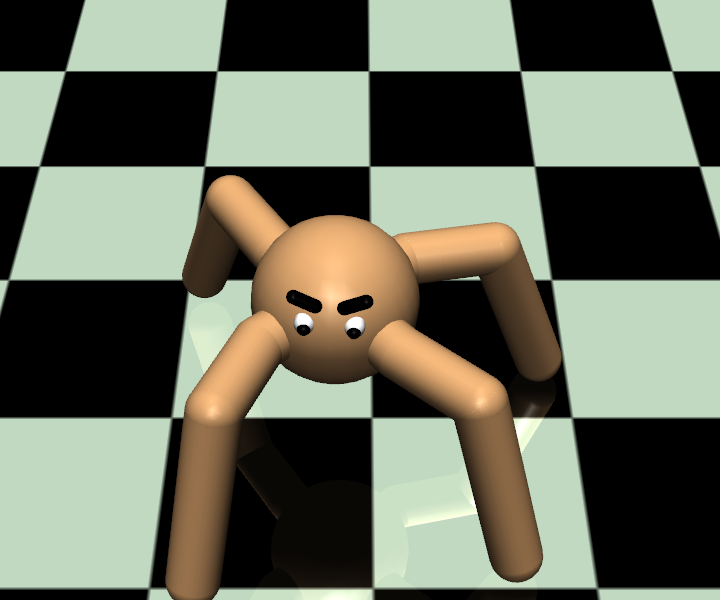}
  \!
    \includegraphics[width=0.2\linewidth]{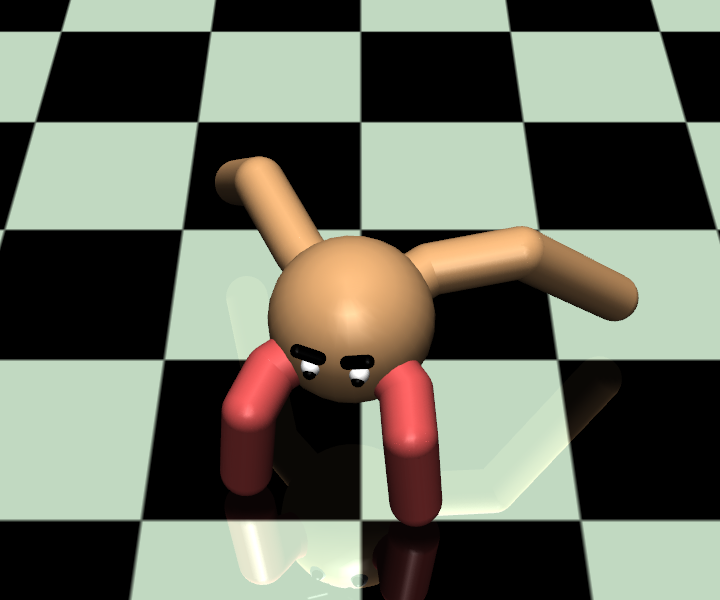}
  \!
    \includegraphics[width=0.2\linewidth]{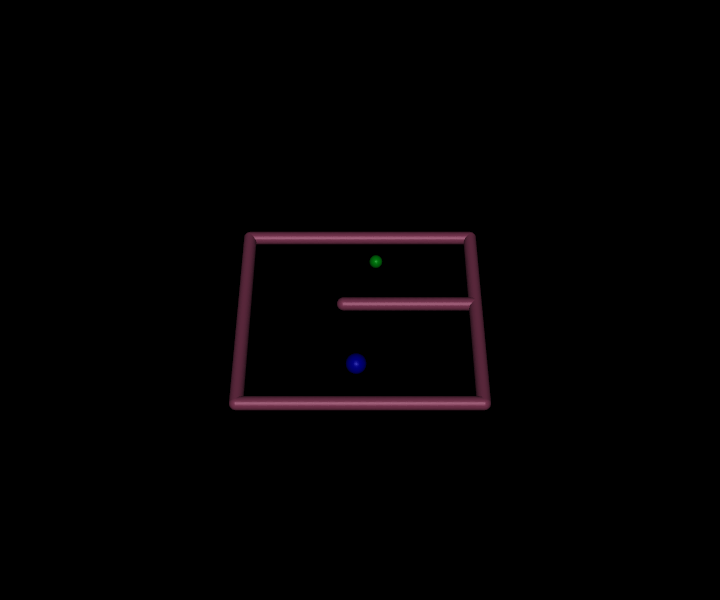}
  \!
    \includegraphics[width=0.2\linewidth]{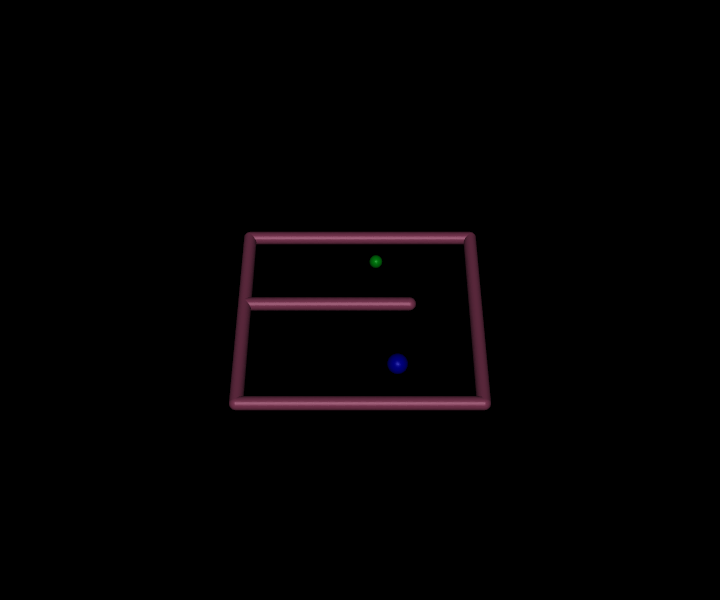}
\caption{ \emph{Top row:} Ant environment is used to perform experiment under criterion 1 \emph{Bottom row:} For shifting maze task, the agent (blue) is required to reach at the goal (green) and is being used for experiment under criterion 2}
\label{transfer_task_envs}
\end{figure}

%%%%%%%%%%%%%%%%%%%%%%%%%%%%%%%%%%%%%%%%%%%%%%%%%%%%%%%%%%%%%%%%%%%%%%%%%%%%%%%%%%%%%%%%%%%%%%%%%%%
\section{Multiplicative Compositional Policies}
%%%%%%%%%%%%%%%%%%%%%%%%%%%%%%%%%%%%%%%%%%%%%%%%%%%%%%%%%%%%%%%%%%%%%%%%%%%%%%%%%%%%%%%%%%%%%%%%%%

Policy often gets overfitted over one task and thus become harder to re-train during transfer learning task. Different techniques such as early stopping or regularization can be helpful in this case. An alternative solution is provided by \cite{VBD} through using \emph{variational bottleneck} technique to control over the information that is fed to the discriminator using mutual information theory, which allows more controlled update in the generator. But this technique is highly dependent on hyper-parameter of the bottleneck itself, thus performance can be varied vastly if not tuned properly for each task. We want to learn a policy that gives a good performance during transfer learning without being dependent on exact hyper-parameter tuning. We propose using \emph{Multiplicative Compositional Policy} (MCP) \cite{MCP} to improve policy performance in transfer learning task. MCP was introduced to combine multiple primitive skills in RL setting and these primitve skills are controlled using a gating function. We propose using multiple primitive networks to learn single skill in IRL setting. Using a weighted multiplicative composition of Gaussian primitives improves imitation performance and allows more flexibility in composing into new policy in transfer learning.

\emph{MCP} \cite{MCP} tries to learn different skills with different primitive policies and then reuse those learned skills by combining them to do a more sophisticated task. In hierarchical learning  \cite{Composablecontrollers, BetweenMDPandSemiMdp, AdditiveExpert, peng2016terrain} it is common to learn premitives and then find a composite policy $\pi(a|s)$ by weighted sum of distribution from premitives $\pi_i(a|s)$. A gating function computes the weight $w_i$, that determines the probability of activating each premitive for a given state, $s$. Composite policy can be written as:
\begin{equation}
    \pi(a|s) = \sum_{i=1}^k w_i(s)\pi_i(a|s) ,  \sum_{i=1}^k w_i(s) = 1, w_i(s) > 0.
\end{equation}
% problem with standard hierarchical setting
Here k denotes the number of primitives and this can be referred as \emph{additive model}. Standard hierarchical learning models can sequentially select particular skill over time. In other words, can activate single primitive at each timestep. \emph{MCP} \cite{MCP} proposes an \emph{multiplicative model}, where multiple primitives are activated at a given time-step. This allows an agent to learn a complex task requiring to perform more than one sub-task at a time.

%\subsubsection{Theory}
MCP \cite{MCP} decomposes agent's policy into premitives, train those primitive policies on different sub-tasks and then obtain a composite-policy by multiplicative composition of these premitives. Here each primitive is considered as distribution over actions and the composite policy is a multiplicative composition of these distribution. Thus,
\begin{equation}
    \pi(a|s,g) = \frac{1}{Z(s,g)} \prod_{i=1}^k \pi_i(a|s,g)^{w_i(s,g)}  , w_i(s,g) \geq 0. 
\end{equation}

MCP enables an agent to activate multiple primitives simultaneously with each primitive specializing in different behaviors that can be composed to produce a continuous spectrum of skills, whereas, standard hierarchical algorithm only activate single primitive at a time.

% ===============================================================================
\section{MCP in off-policy-AIRL}
% ===============================================================================

% contribution: 
MCP learns premitive policies $\pi_i$ by training them into different sub-tasks and then try to learn more complex task leveraging the learned skills. But in transfer learning experiments, we use MCP to learn single task. The underlying idea is to observe whether it can relearn the composition of different action parameters when we put them in dynamic scenarios. For example, during transfer learning experiments \cite{AIRLfu2018} disables two legs and thus it is important for the agent to re-optimise it's policy to walk in this transfer learning setup. Our initial objective is to see if MCP can decompose motor skills for a single task and re-optimise the learned policy by re-training the gating function. We refer this new policy as SAC-MCP. %Will try only updating gating function (with or without prior gating) to move forward.

In MCP \cite{MCP}, experiments were conducted using on-policy algorithms in RL setting and the policy parameters were considered to be state and goal dependent. We explore this concept to decompose motor skill over single task in IRL setting using off-policy algorithm. 
% ==============
% figure: SAC_MCP actor network
% ==============
\begin{figure}[hbt!]
\centering
   \includegraphics[width=0.6\linewidth]{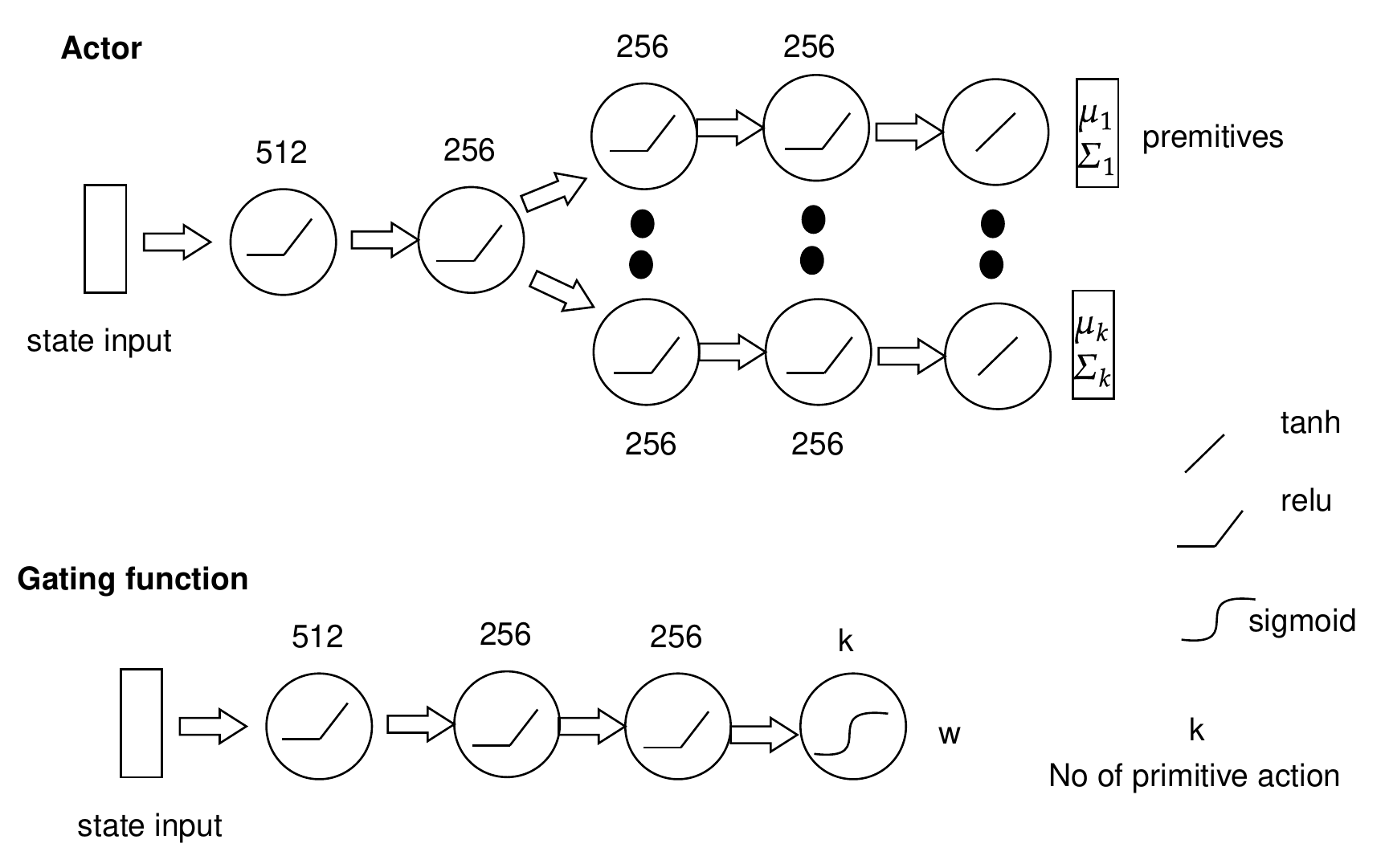}
\caption{Generator architecture for SAC-MCP.}
\label{SAC_MCP_ACTOR}
\end{figure}

% %============================
% ===================================
\section{Experimental Setup}
\label{experiment setup}
% ===================================

% collect expert may be added at appendix
% To compute imitation learning and IRL, learner needs expert demonstrations. We have compared the performance on TD3, SAC in RL setting for our expert selection. We have trained policy on MuJoCo tasks in RL setting to collect expert data for imitation performance comparison and also collected expert trajectory for environments used in AIRL \cite{AIRLfu2018} to compare transfer learning in following chapters. 

Similar to \cite{GAILHoE16,DACkostrikov2018} discriminator is 2 layer MLP of 100 hidden units with tanh activation. Our generator consists of separate Actor and Critic neural network and follows the architecture used in \cite{DACkostrikov2018,TD3}, where both of these networks have 2 layer MLP of 400 and 300 hidden units with ReLU activation. To implement SAC-MCP, we modify Actor network (see Figure \ref{SAC_MCP_ACTOR}) by adding multiple premitive networks with same configuration described in \cite{MCP}. 

We have trained all networks with the Adam optimizer \cite{adam} from \emph{PyTorch}, which uses default learning rate of $1e^{-3}$. For our experiment we trained our algorithms on \emph{MuJoCo} \cite{mujocotodorov2012} continuous control task environments. For transfer learning experiments we use Custom-Ant and Shifting Maze environments from \cite{AIRLfu2018}. Performance curve is obtained using the mean over 10 experiments for 0-9 seeds and evaluated after each 5000 interaction with the environment. During each evaluation we have stored the average performance of 10 runs.
 
% %============================

%=================
% =========================
\section{Results}
% =========================
\label{results}

\subsection{Reward function selection}
% ========================================

\begin{figure}[hbt!]
\centering
\includegraphics[width=0.48\linewidth]{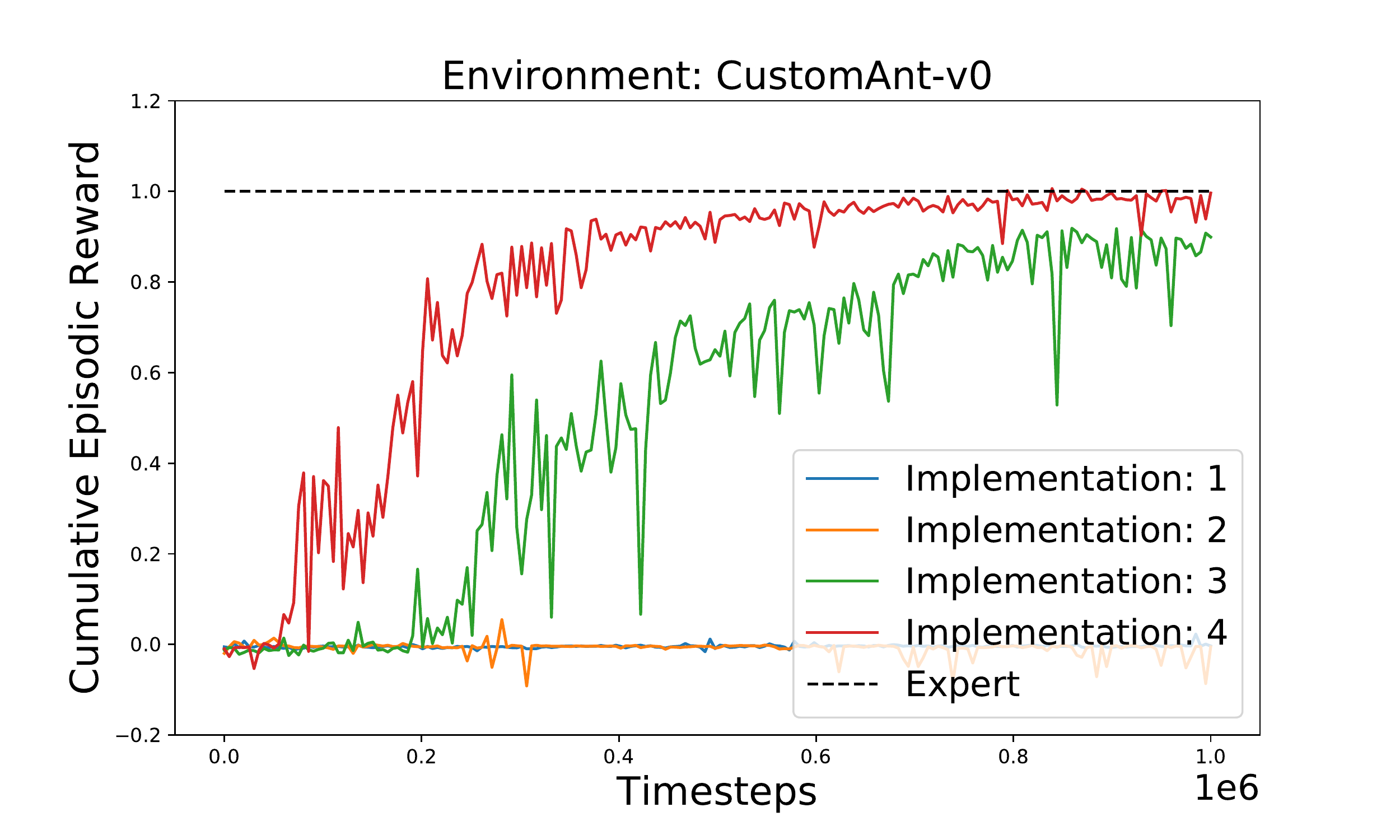}
\!
\includegraphics[width=0.48\linewidth]{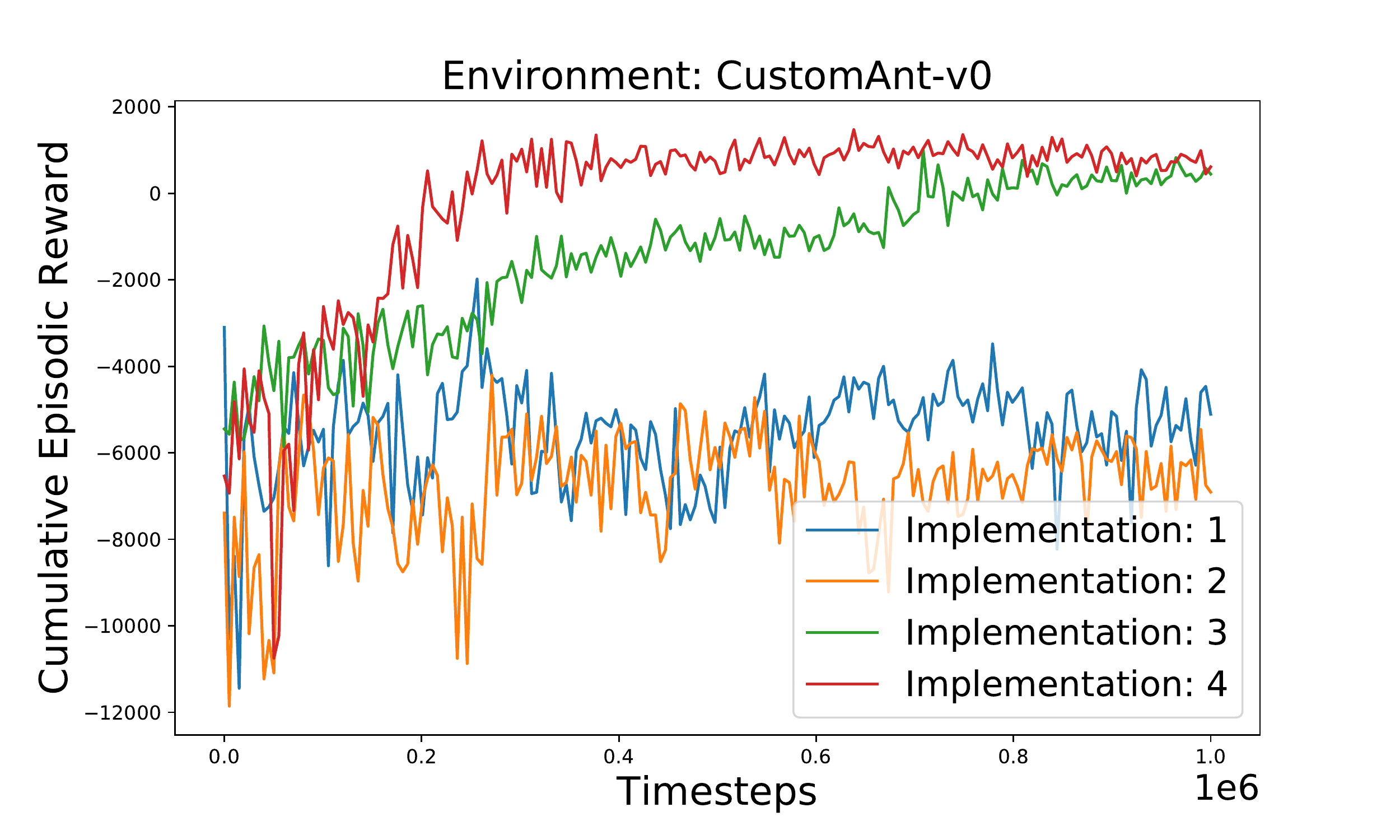}

\caption{Performance comparison of Off-policy AIRL for different update rules. Here figure (a) gives the performance curve using actual reward from the environment (b) shows the cumulative reward achieved from reward approximator $r_\theta$}
\label{implementation4}
\end{figure}

%%%%%%%%%%%%%%%%%%%%%%%%%%%%%%%%%%%%%%%%%%%%%%%%%%%%%%%%%%%%%%%%%%%%%%%%
We conduct experiment on \emph{CustomAnt-v0} \cite{AIRLfu2018} to select reward function for policy update. We train SAC policy and collected 50 trajectory from the expert. We keep our seed fixed to $0$. Figure \ref{implementation4}$(a)$ gives the performance curve using actual reward from the environment and it is evident that reward signal directly from the reward function approximator (Implementation 4) gives better policy performance in imitation task. We also see the cumulative reward achieved from reward approximator $r_\theta$ (see Figure \ref{implementation4}$(b)$) replicates the actual performance curve. As our experiment shows drastic difference in performance we have not done comparative study for multiple seeds or other environments.

% ======
% figure: implementation
% ======

\begin{figure}[hbt!]
\centering
   \includegraphics[width=0.48\linewidth]{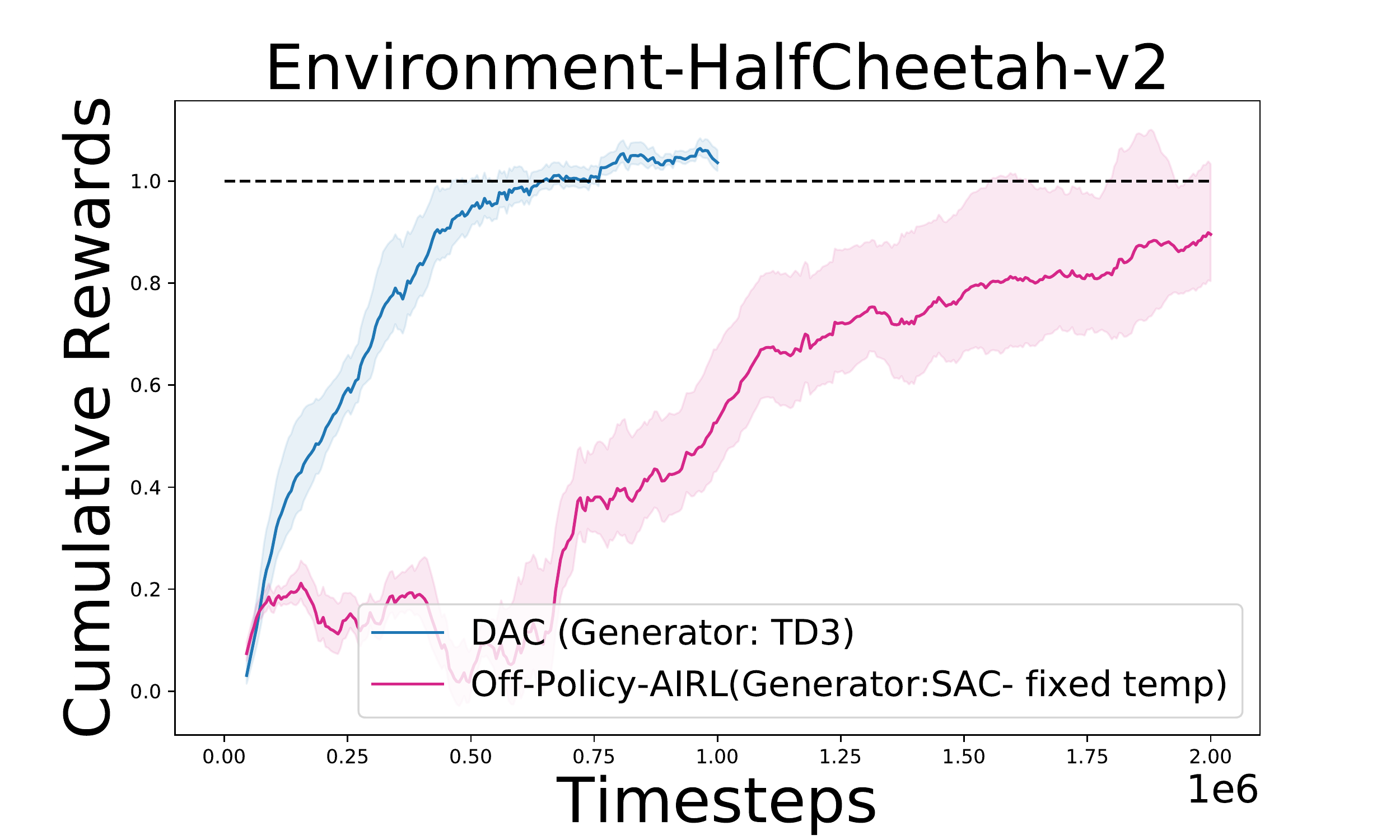}
  \!
    \includegraphics[width=0.48\linewidth]{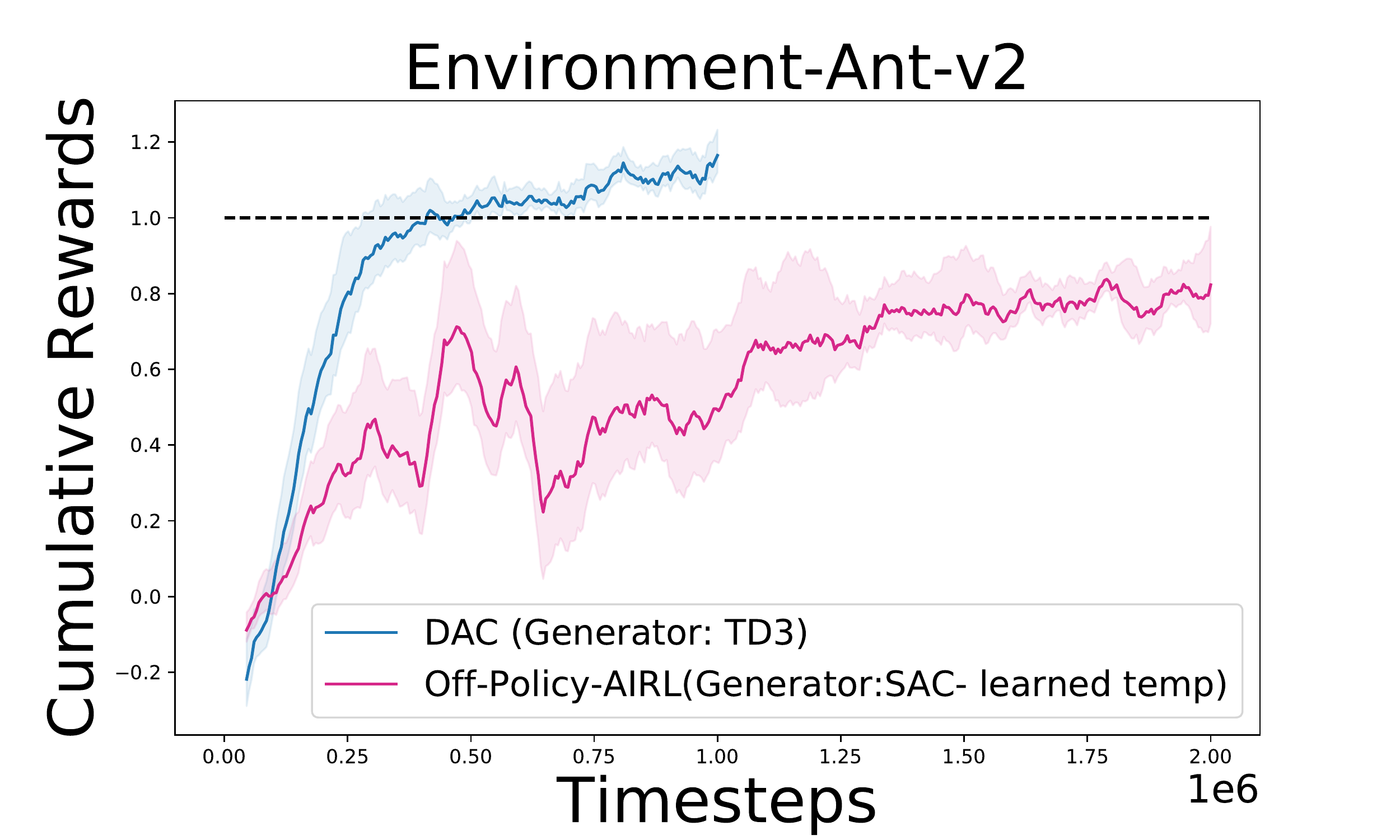}
  \!
    \includegraphics[width=0.48\linewidth]{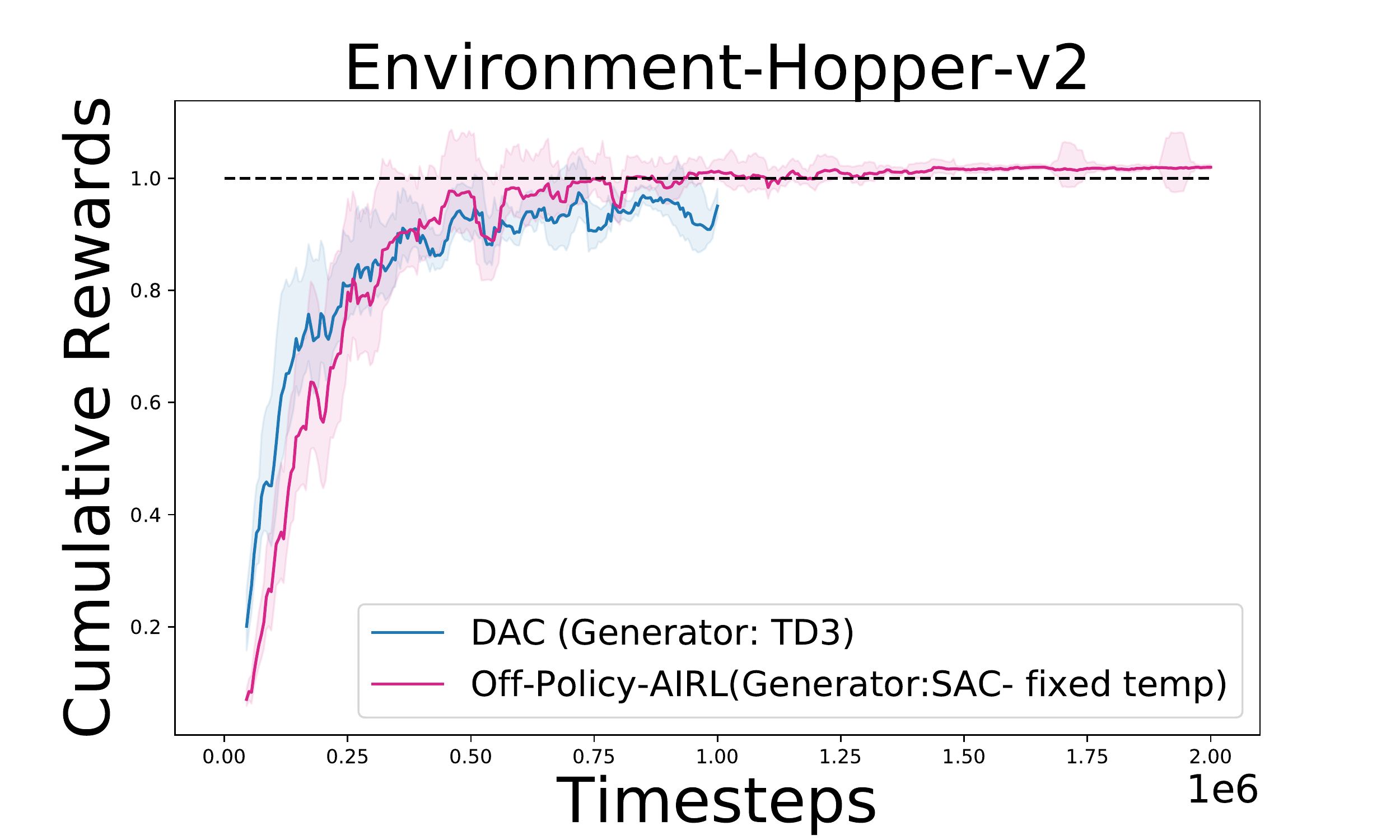}
  \!
    \includegraphics[width=0.48\linewidth]{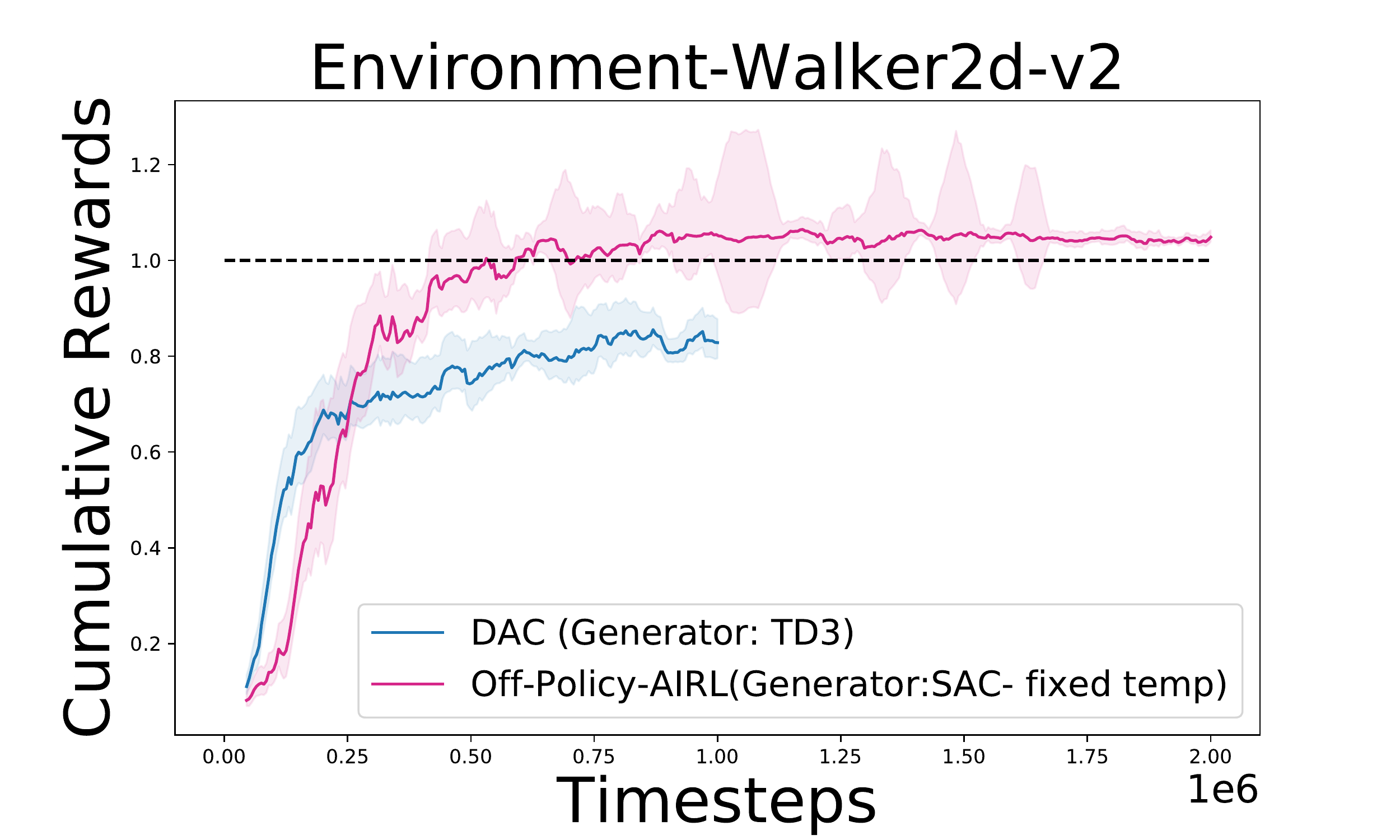}
\caption{Performance Comparison of DAC and Off-policy AIRL over continuous control task}
\label{OffPolicyAIRL_SAC}
\end{figure}

% ========================================
As demonstrated in Figure \ref{OffPolicyAIRL_SAC} using implementation-4 off-policy AIRL gives a better imitation performance than DAC in Hopper-v2 and Walker2d-v2 environments within $1e^6$ iterations, while provides comparable performance for HalfCheetah-v2 and Ant-v2 environments after $2e^6$ iterations. 

\subsection{Performance in transfer learning}

% ======
% figure: off-policy AIRL SAC
% ======

We test trained agent in transfer learning task under criterion (1) and (2). We demonstrate the utility of learning reward function which gives us an advantage of learning IRL over imitation learning. In IRL setting, Off-policy AIRL learns expert like behavior in Custom Ant environment but suffers from variance in Shifting maze task. We do not tune separate hyper-parameter for individual environment but doing so may reduce performance variance for Shifting maze environment.

% ==== figure Generator: SAC

\begin{figure}[hbt!]
\centering

\includegraphics[width=0.48\linewidth]{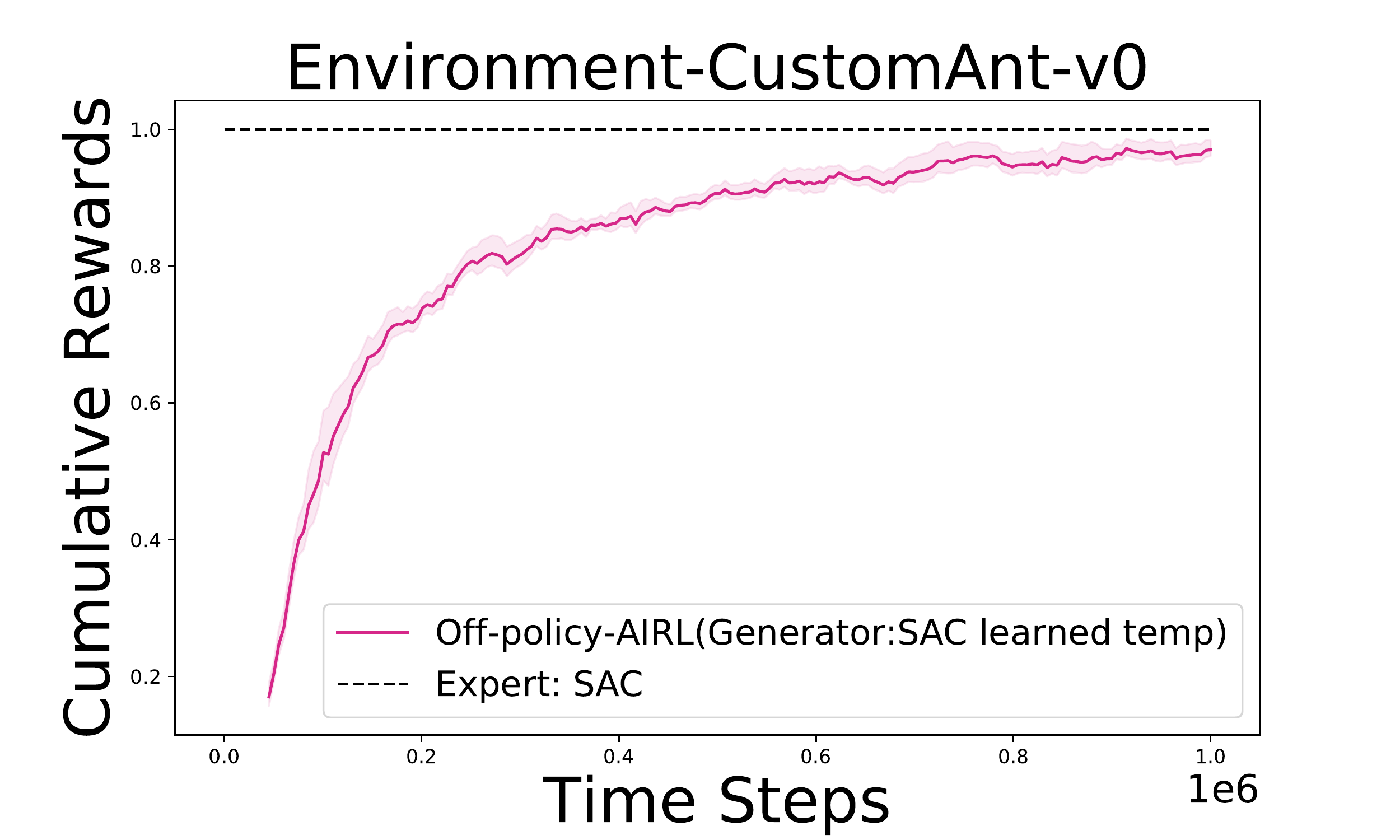}
\!
\includegraphics[width=0.48\linewidth]{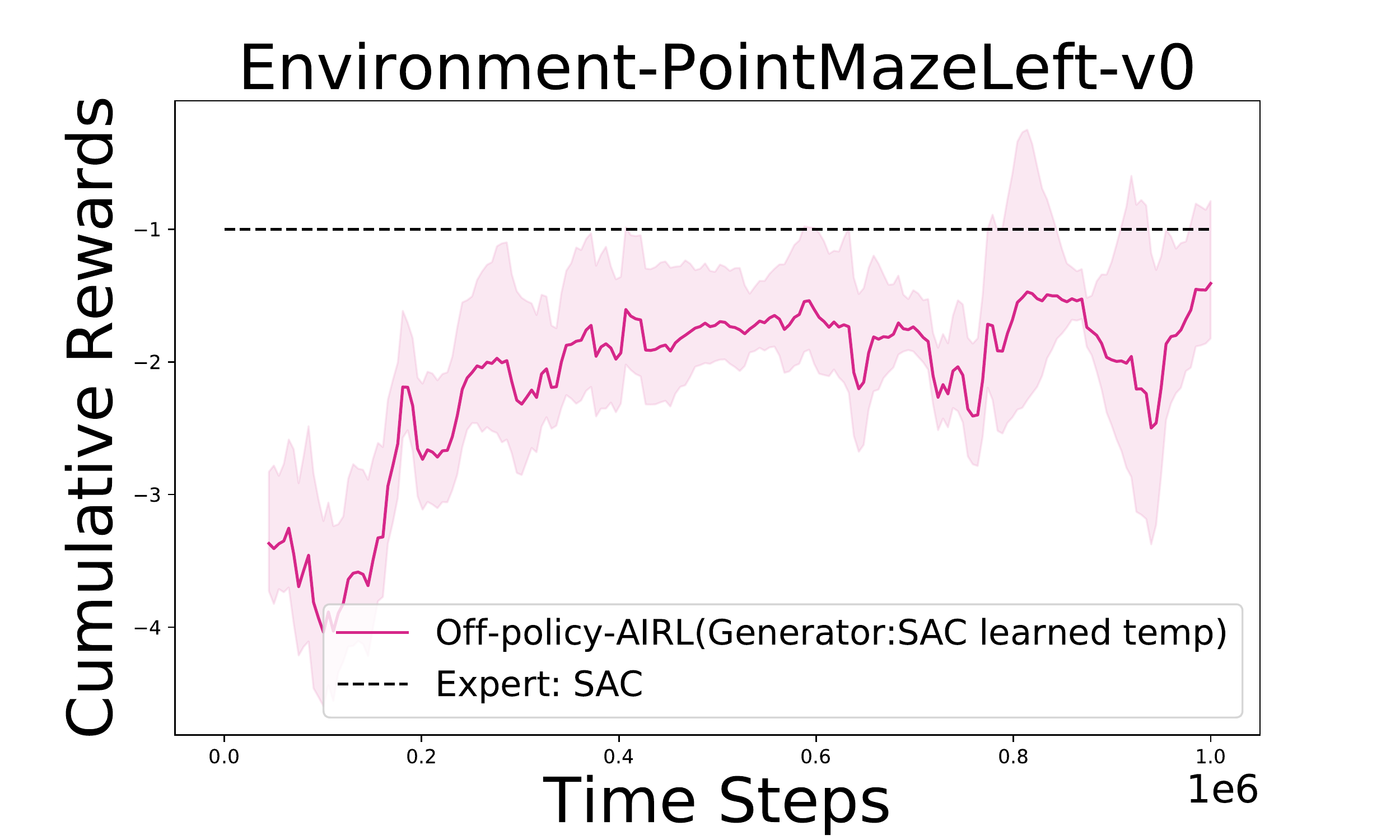}
\!
\includegraphics[width=0.48\linewidth]{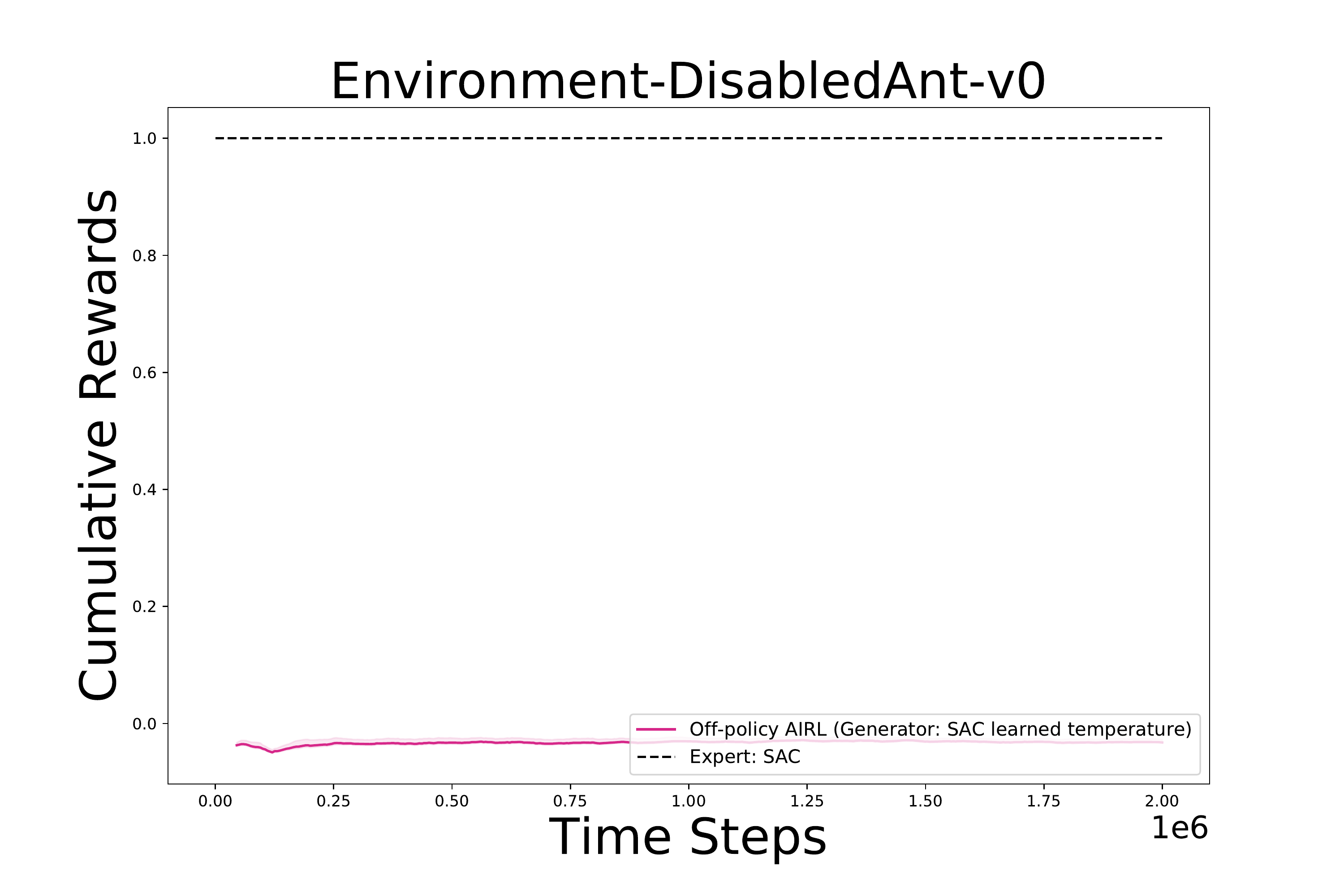}
\!
\includegraphics[width=0.48\linewidth]{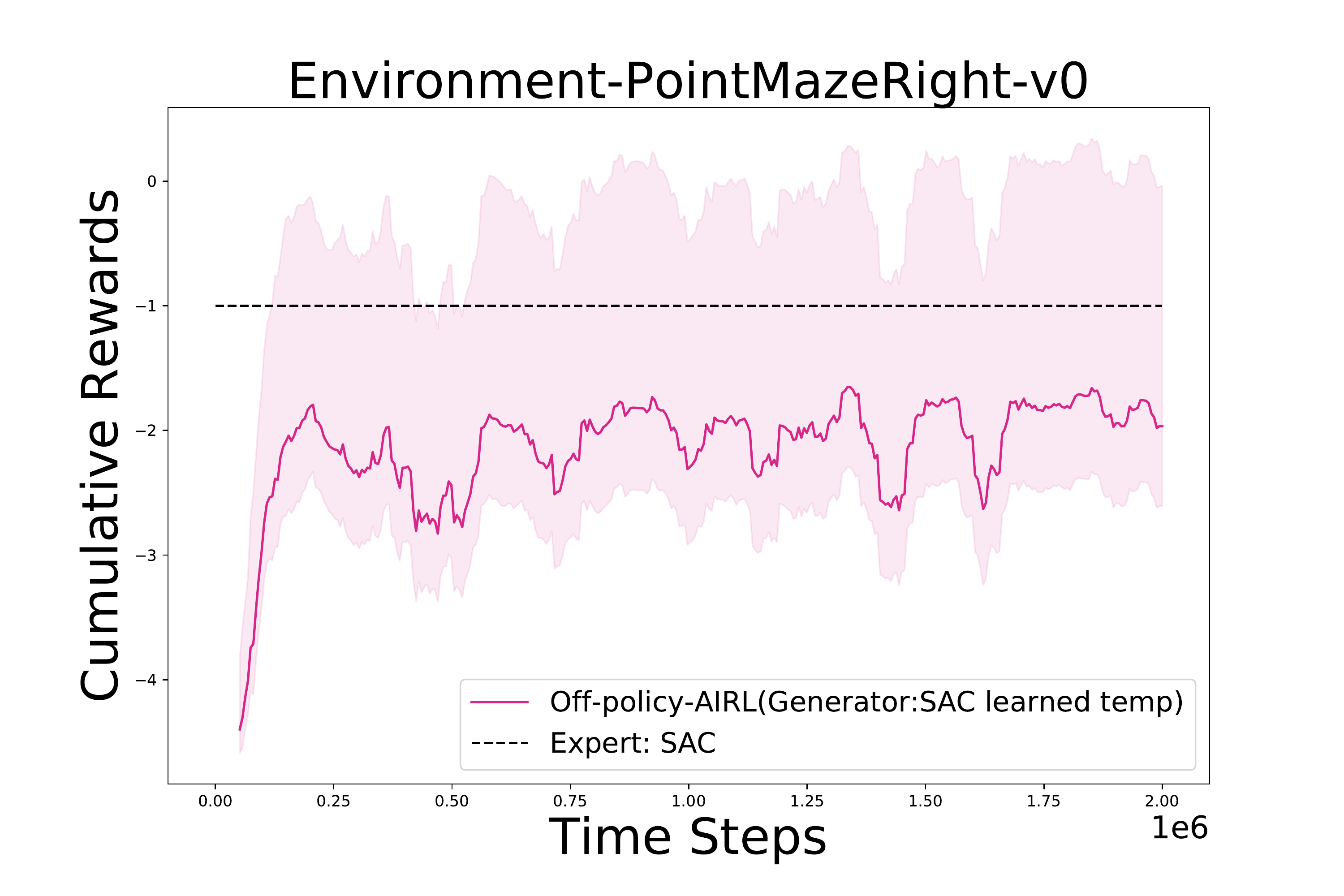}

\caption{\emph{Top row:}Performance Off-policy AIRL over continuous control task  \emph{Bottom row:}Performance Off-policy AIRL during transfer learning task.}
\label{OffPolicyAIRL_SAC_transfer}
\end{figure}

As illustrated in Figure \ref{transfer_task_envs} during transfer learning under criterion-1, agent's gait changes significantly and as can be seen in Figure \ref{OffPolicyAIRL_SAC} off-policy-AIRL fails to re-learn the policy using reward function. But we do not conclude that it is not possible to relearn policy under this criterion using off-policy-AIRL. It is demonstrated in \cite{Ha2018designrl} that changing a better structure of agent's body not only is better suited for the task but also facilitates policy learning. The environment that we use in criterion-1 was introduced in \cite{AIRLfu2018}, where the structure of the body of the quadrupedal Ant was changed \cite{AIRLfu2018} for transfer learning by shrinking two legs and making other two larger. But the specific increment or decrement of the legs may have favored for \cite{AIRLfu2018} to re-learn policy when two legs are paralyzed in transfer learning. A hyper-parameter tuning of the structural change may work for our proposed algorithm as well. Experiments with more structural flexibility are required to come to a solid conclusion under this criterion.

On the other hand for Shifting maze environment (see Figure \ref{OffPolicyAIRL_SAC}) it successfully completes the task but suffer from high variance in performance.

% ===========================================
\subsection{Performance comparison for using MCP}

\begin{figure}[hbt!]
\begin{center}
    \includegraphics[width=0.48\linewidth]{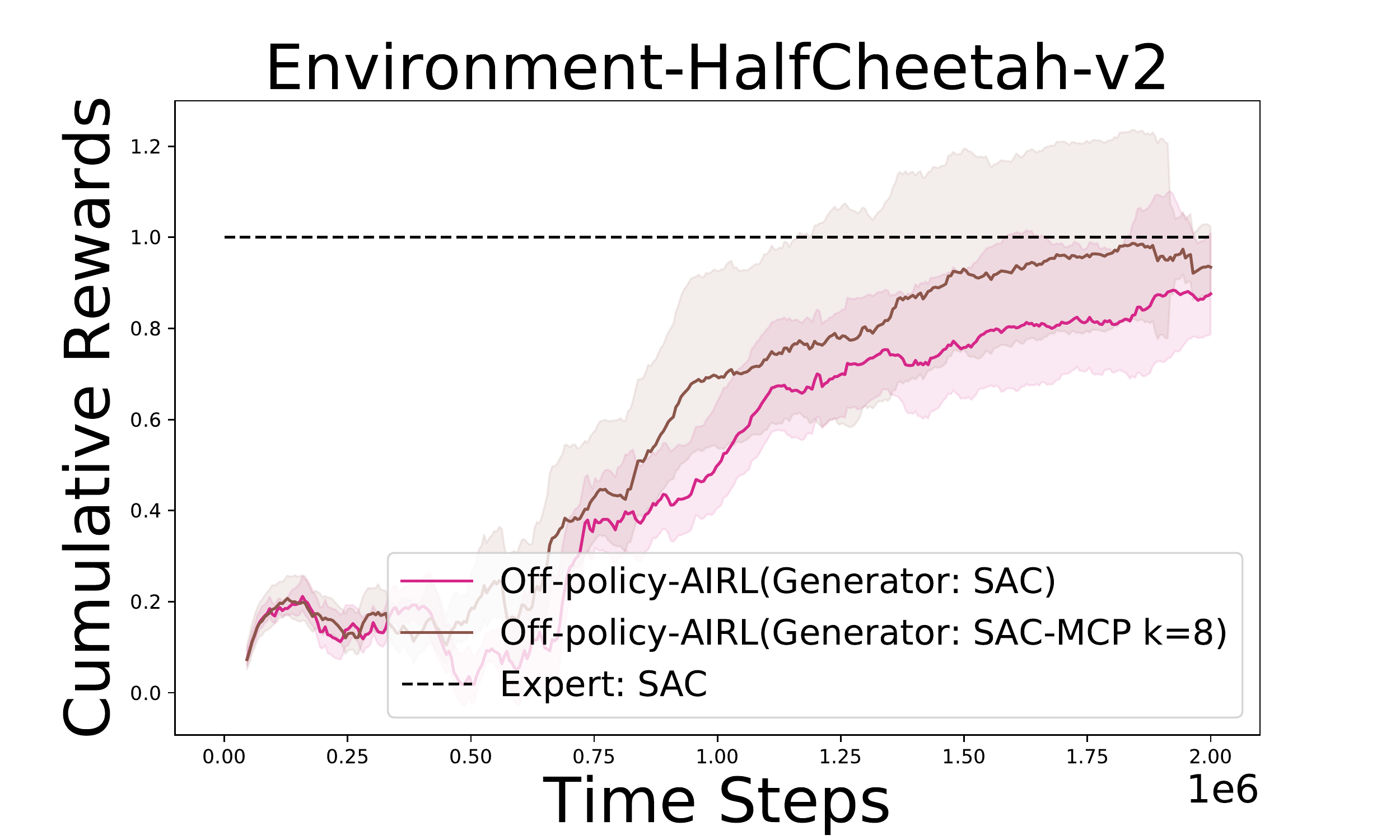}
   \!
    \includegraphics[width=0.48\linewidth]{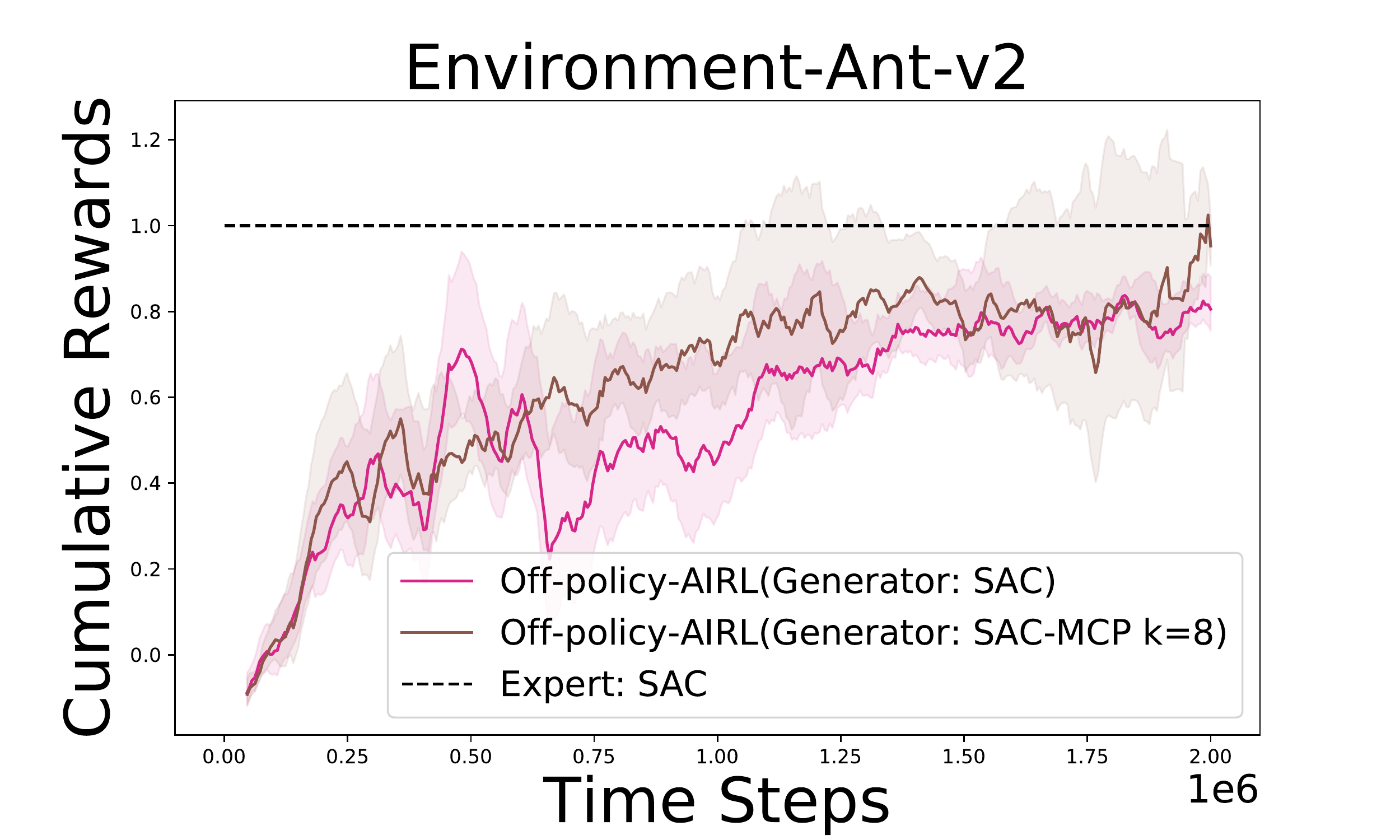}
   \!
    \includegraphics[width=0.48\linewidth]{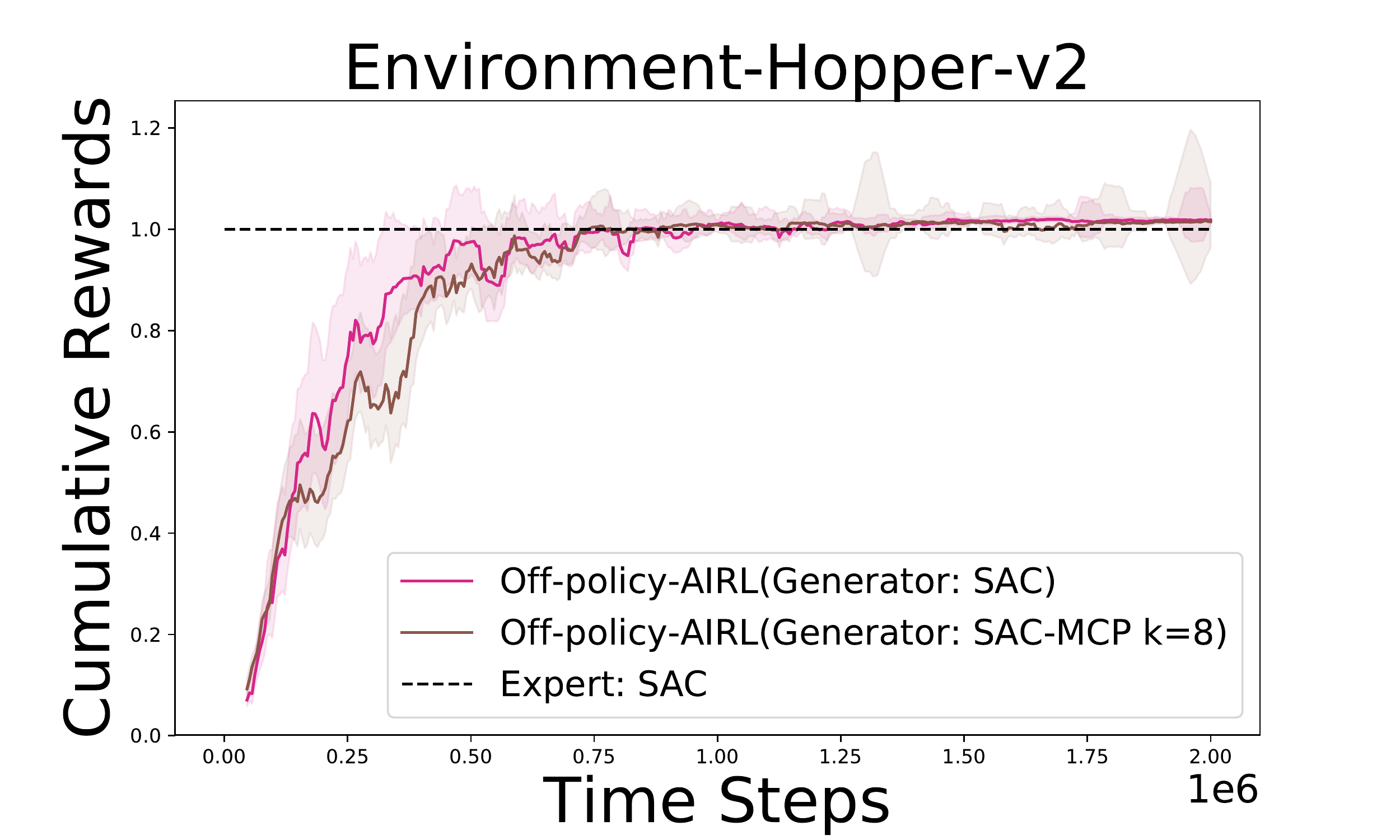}
   \!
    \includegraphics[width=0.48\linewidth]{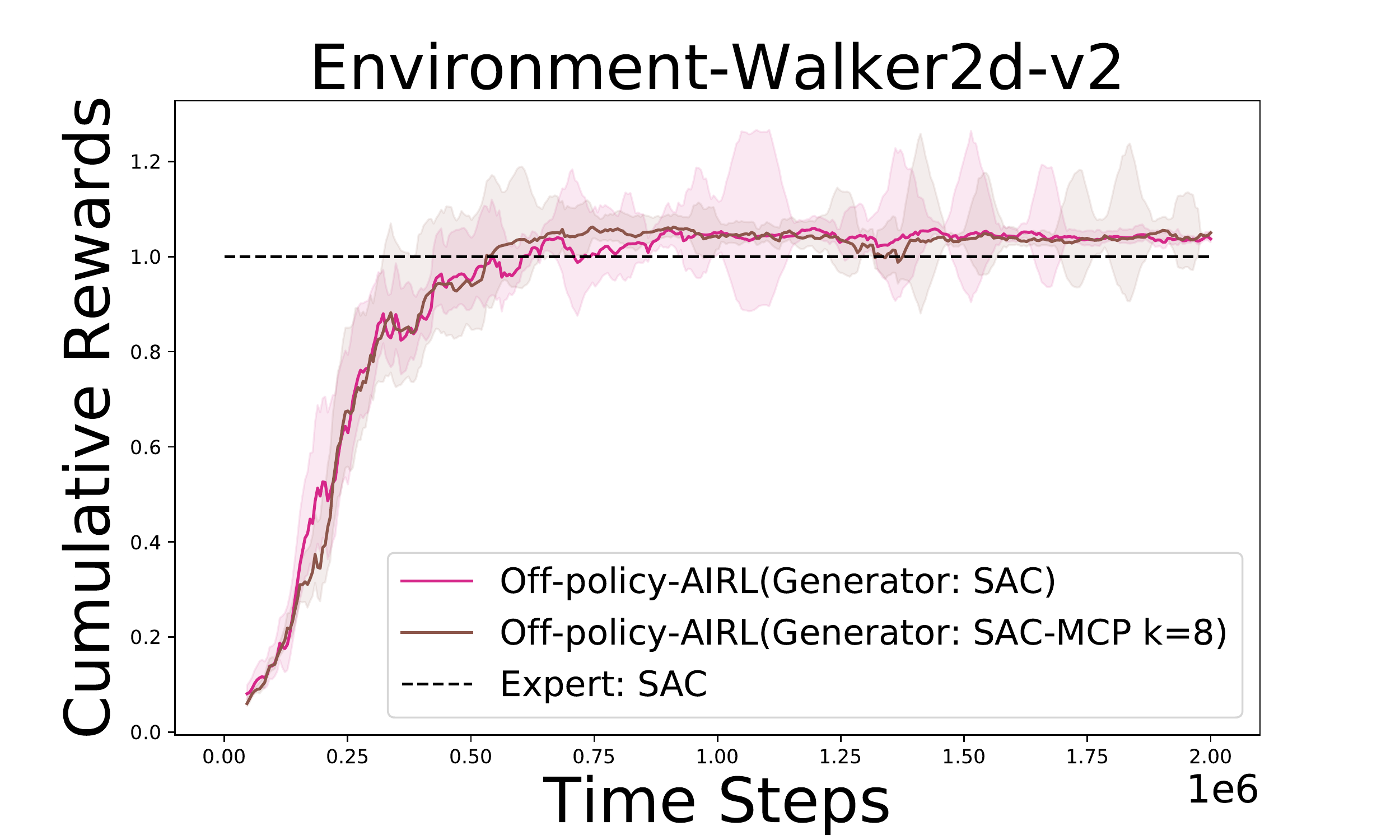}
\caption{Performance Comparison of SAC and SAC-MCP as generator in off-policy-AIRL over continuous control task.}
\label{OffPolicyAIRL_SAC_MCP}
\end{center}
\end{figure}

Performance of our MCP implementation in IRL setting is shown in Figure \ref{OffPolicyAIRL_SAC_MCP}. Using SAC-MCP improves the imitation performance in multiple (HalfCheetah-v2 and Ant-v2) MuJoCo control task.

% ======
% figure: off-policy-AIRL- SAC vs SAC_MCP
% =====

\begin{figure}[hbt!]
\centering
   \includegraphics[width=0.48\linewidth]{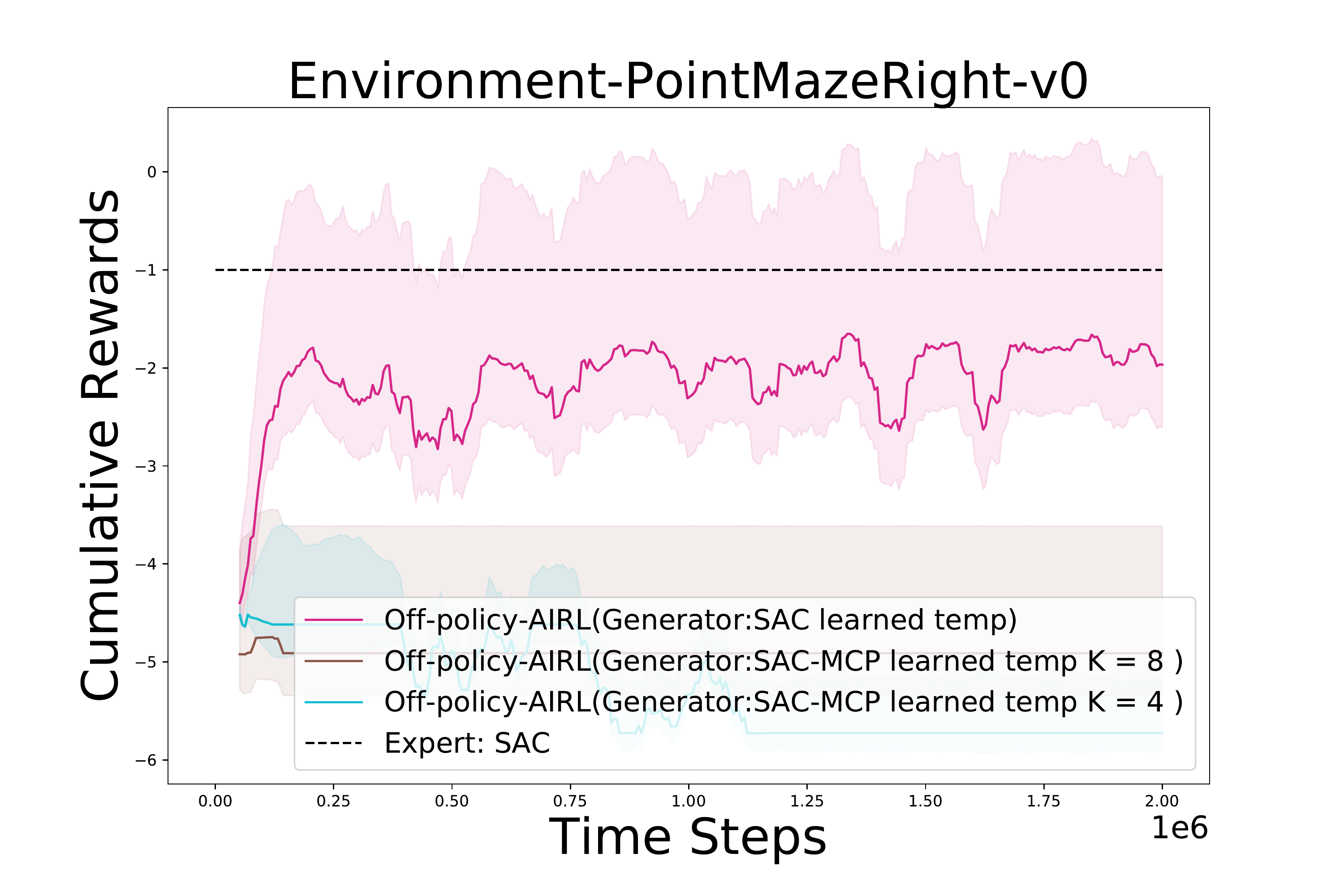}
   \!
    \includegraphics[width=0.48\linewidth]{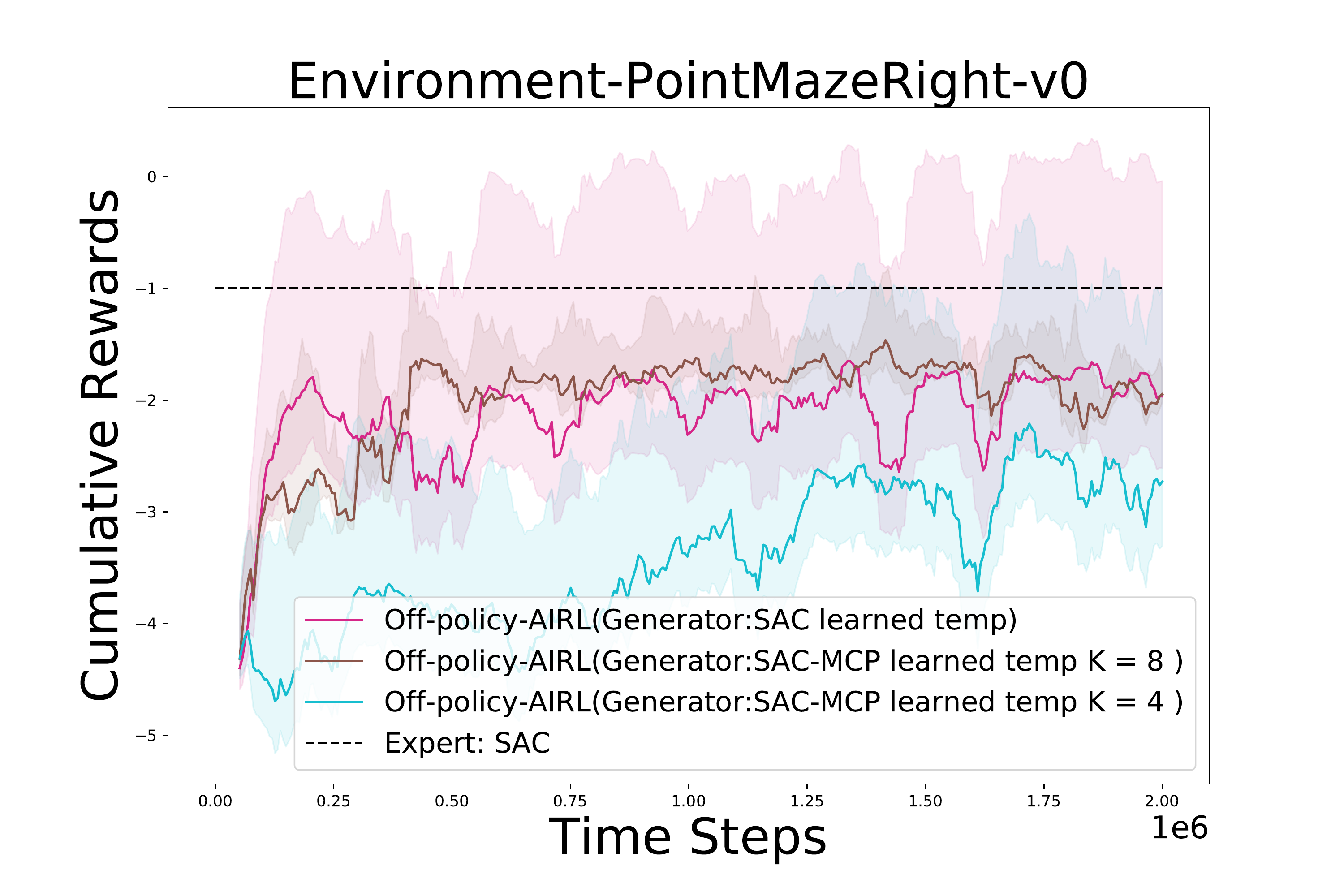}
\caption{Performance of off-policy-AIRL (Generator: SAC-MCP) during transfer learning. \emph{Left}:Train only gating function. \emph{Right}:Train both policy and gating function}
\label{TransferTask_SACMCP}
\end{figure}

We again conduct our transfer learning experiments on Shifting-Maze tasks. Here we do not load prior gating function. We compare performance with and without re-training the actor. Our initial hypothesis is that having multiplicative composition of Gaussian primitives should allow us with diverse behavior by only training the gating function during transfer task. But Figure \ref{TransferTask_SACMCP} shows we are not able to learn shifting maze task without retraining the actor networks. We compare performance of SAC-MCP for $k=\{4,8\}$ premitives (see Figure \ref{TransferTask_SACMCP}). For $K=8$ it reduces the performance variance suffered by SAC.

% ====

% ====

\subsection{Random State Initialization}

\begin{figure}[hbt!]
\centering
   \includegraphics[width=0.48\linewidth]{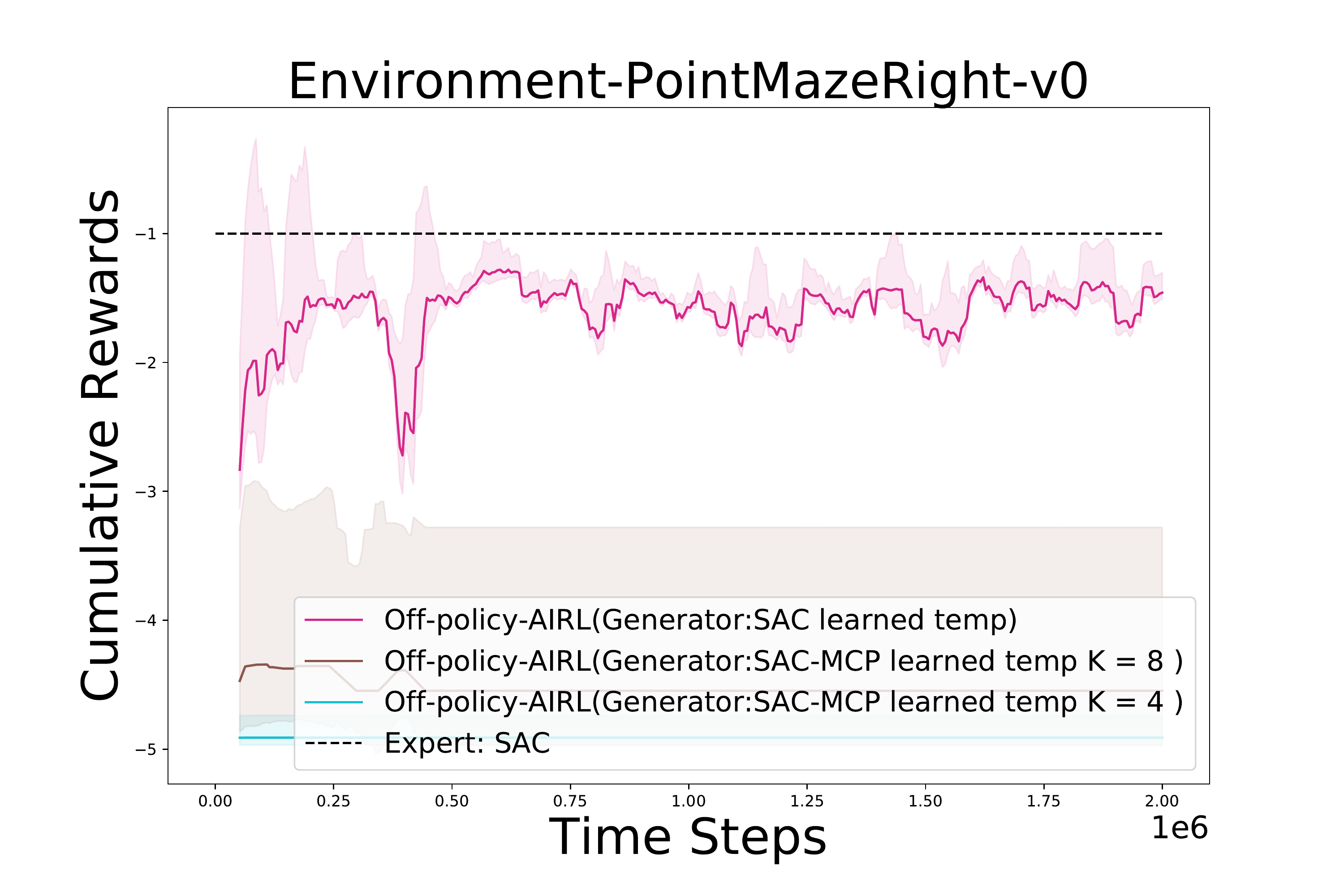}
   \!
    \includegraphics[width=0.48\linewidth]{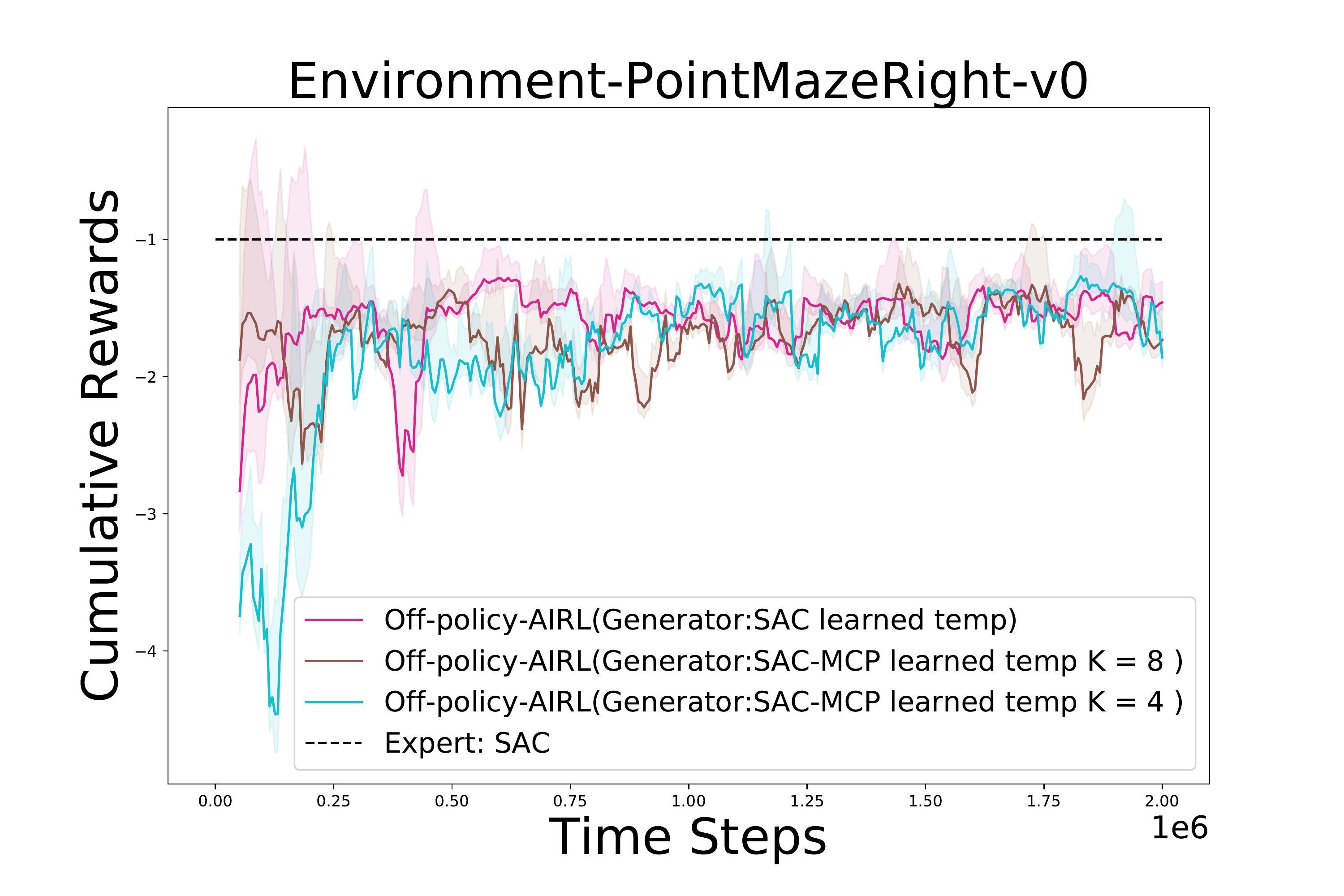}
\caption{Performance Comparison of SAC and SAC-MCP as generator during transfer learning with random state initialization. \emph{Left}:Train only gating function of SAC-MCP. \emph{Right}:Train both policy and gating function of SAC-MCP}
\label{TransferTask_SACMCP_random_init}
\end{figure}

We initialize the environment from a random state. As we retrain already learned policy over one task, it does not give much exploratory action. Thus initializing environment at different state will help to explore the new task. We use expert agent to run for 0-100 steps and then stop at random, thus enabling the learning agent to start at random state every time the environment resets. It allows an agent to explore from a state that is expected to be visited by an expert. At the same time, using learned policy (from prior imitation task) to take actions will result into better quality of sequential observations. We see in Figure \ref{TransferTask_SACMCP_random_init} random initialization improves performance and also reduces performance variance for both SAC and SAC-MCP on transfer learning task for shifting maze environment.

\newpage
%=================

\newpage
%===================
\section{Discussion and Conclusion}
\label{conclusion}
We propose an off-policy IRL algorithm in adversarial learning which learns policy using its approximated reward function. We use SAC as generator to improve imitation performance while still being sample efficient. We compare different way to compute reward functions for off-policy generator update and experimentally find simple state dependent reward network without entropy regularization works best for our algorithm. We compare our imitation performance with state-of-the-art imitation learning algorithm to show our persistent expert like performance over continuous control environments.

We experimentally demonstrate the utility of learned reward function using off-policy-AIRL when the expert demonstrations are non-existent in transfer learning. We compare performance over two transfer learning criterion, where either (1) the dynamics of the agent or (2) the dynamics of the environment changes while the goal/objective and action dimension remain the same.  

Furthermore, we propose using MCP model to learn single primitive to achieve more flexibility in retraining policy in transfer learning. We observe an improved performance for using SAC-MCP on both imitation performance and transfer learning. We also conduct experiments under random state initialization, which helps the agent to explore from an state that is expected to be visited in the new environment and reduces performance variance over prior experiments.

But unfortunately, our algorithm (using both SAC and SAC-MCP as generator) fails when the (criterion 1) dynamics of the agent is changed in transfer learning. The quadrupedal Ant has a very specific increment and decrement of the legs, which may have favored for \cite{AIRLfu2018} to re-learn policy when two legs are paralyzed in transfer learning. As specific body structure of an agent can facilitate learning policy over a task \cite{Ha2018designrl}, experiments using environments with more structural flexibility is required to come to a conclusion under this criterion.
%===================

%=========================
\section{Acknowledgements} 

The author thank \emph{Aditya Mahajan} for reviewing the paper and providing valuable feedback. Implementation of \emph{Discriminator Actor-Critic} algorithm used in this paper is initially implemented as ICLR-2019 Reproducibility Challenge along with \emph{Sheldon Benard} and \emph{Vincent Luczkow} \cite{DAC_ICLR_REPRODUCE}. The author also thank \emph{Ilya Kostrikov} for looking into the code and helping further on the implementation. The code that is used in this paper is an extension of the project. Experiments and respective results mentioned are computed for the purpose of this paper.  
%=========================

\bibliographystyle{unsrt}  
\bibliography{references}  %%% Remove comment to use the external .bib file (using bibtex).

%%% Comment out this section when you 

%\bibliography{references} is enabled.

\newpage
%===================
\section{Appendix: Off-Policy Adversarial Inverse Reinforcement Learning}

%========================================================================
\subsection{Off-policy-AIRL algorithm}
% % ===============================
% % Algorithm
% % ===============================
% \begin{algorithm}
% \footnotesize
% \caption{Off-policy AIRL Algorithm}\label{alg:dac}
% \begin{algorithmic}[1]

% %\Procedure{Euclid}{$a,b$}\Comment{The g.c.d. of a and b}
% \STATE \textbf{Input}: Expert Replay Buffer $\mathcal{R_E} $ 

% \WHILE{$n\not=1e^6$}%\Comment{We have the answer if r is 0} 
%     \STATE episodic timestep = 0
% \WHILE{$not \ done$} 
%     \STATE $a \gets{\pi(a|s)}$
%     \STATE $\mathcal{R}\gets \mathcal{R} \cup {s,a,.,s'}$
%     \STATE episodic timestep += 1
%     \STATE n += 1
%     \STATE \textbf{if}  $done$ WarpAbsorbingState 
% \ENDWHILE

% \FOR{$episodic \ timesteps$}
%     \STATE sample random mini-batch: { $ \{ (s_b, a_b, ., s_b^{'}) \}_{b=1}^B \in \mathcal{R}$ }
%     \STATE sample random mini-batch: { $ \{ (s_{b_E}, a_{b_E}, ., s_{b_E}^{'}) \}_{b=1}^B \in \mathcal{R_E}$}
%     \STATE Compute loss:{ $\mathcal{L}_{\theta \phi} = \sum_{b=1}^B [- log D_{\theta \phi}(s_b,a_b,s_b^{'}) -  log (1-D_{\theta \phi}(s_{b_E},a_{b_E},s_{b_E}^{'} ))]$ }  
%     \STATE update $D_{\theta \phi}$ + gradient-penalty
% \ENDFOR

% \FOR{$episodic \ timesteps$}
%     \STATE sample random mini-batch {$ \{ (s_b, a_b, ., s_b^{'} ) \}_{b=1}^B \in \mathcal{R}$}
%     \STATE use current reward function { $\{ (s_b, a_b, r_b, s_b^{'}) \}_{b=1}^B \leftarrow r_\theta(s_b)$}
%     \STATE update SAC
% \ENDFOR
% \ENDWHILE
% \end{algorithmic}
% \end{algorithm}

% ===============================
% Algorithm
% ===============================
\begin{algorithm}
\caption{off-policy-AIRL Algorithm}\label{alg:dac}
\begin{algorithmic}[1]

%\Procedure{Euclid}{$a,b$}\Comment{The g.c.d. of a and b}
\State \textbf{Input}: Expert Replay Buffer $\mathcal{R_E}$  

% \While{$n\not=10^6$}\Comment{We have the answer if r is 0} 
\While{$n\not=10^6$}
    \State episodic timestep = 0
\While{$not \ done$} 
    \State $a \gets{\pi(a|s)}$
    \State $\mathcal{R}\gets \mathcal{R} \cup {s,a,.,s'}$
    \State episodic timestep += 1
    \State n += 1
    \State \textbf{if}  \emph{done} WarpAbsorbingState 
\EndWhile

\For{\texttt{episodic timesteps}}
    \State sample random mini-batch $ \{ (s_b, a_b, ., s_b^{'}) \}_{b=1}^B \in \mathcal{R}$
    \State sample random mini-batch $ \{ (s_{b_E}, a_{b_E}, ., s_{b_E}^{'}) \}_{b=1}^B \in \mathcal{R_E}$
    \State Compute loss: $\mathcal{L}_{\psi,\omega} = \sum_{b=1}^B [- \log D_{\psi,\omega}(s_b,a_b,s_b^{'}) - \log (1-D_{\psi,\omega}(s_{b_E},a_{b_E},s_{b_E}^{'} ))]$   
    \State update $D_{\psi,\omega}$ + gradient-penalty
\EndFor

\For{\texttt{episodic timesteps}}
    \State sample random mini-batch $ \{ (s_b, a_b, ., s_b^{'} ) \}_{b=1}^B \in \mathcal{R}$
    \State use current reward function $\{ (s_b, a_b, r_b, s_b^{'}) \}_{b=1}^B \leftarrow r_\psi(s_b)$
    \State update SAC
\EndFor
\EndWhile

% \EndWhile
% \While{}
% \State $a\gets b$
% \State $b\gets r$
% \State $r\gets a\bmod b$
% \EndWhile\label{euclidendwhile}
%\State \textbf{return} $b$\Comment{The gcd is b}
%\EndProcedure

\end{algorithmic}
\end{algorithm}

% ===================================
\subsection{More Experimental Details}
% ===================================

% \subsubsection{Experimental Setup}
% \label{experiment setup}

% Similar to \cite{GAILHoE16,DACkostrikov2018} discriminator is 2 layer MLP of 100 hidden units with tanh activation. Our generator consists of separate Actor and Critic neural network and follows the architecture used in \cite{DACkostrikov2018,TD3}, where both of these networks have 2 layer MLP of 400 and 300 hidden units with ReLU activation. At the beginning of the training generator takes random exploratory actions, within a few iterations of the discriminator update, it easily classifies the expert from the generator data and overfits on the training data \cite{weight_clipping,gp}. Thus similar to DAC \cite{DAC}, to make the learning more stable we use \emph{gradient penalty} \cite{gp}.

% We have trained all networks with the Adam optimizer \cite{adam} from \emph{PyTorch}, which uses default learning rate of $1e^{-3}$. For our experiment we trained our algorithms on \emph{MuJoCo} \cite{mujocotodorov2012} continuous control task environments. For transfer learning experiments we're using Ant and Shifting Maze environments from \cite{AIRLfu2018}. Performance curve is obtained using the mean over 10 experiments for 0-9 seeds and evaluated after each 5000 interaction with the environment. During each evaluation we have stored the average performance of 10 runs.

%==============================================
\subsubsection{Expert Data Collection}
% ==============================================

%collect expert trajectory
We require expert demonstrations to train policy in imitation task. So we train TD3 \cite{TD3}, SAC\cite{SAC} to compare the performance. To collect the expert data we run experiments on MuJoCo continuous control task and also on environments used in \cite{AIRLfu2018} in RL setting (consider reward function available from the environment) for 1 million iterations. The \emph{temperature parameter} $\alpha$ controls the stochasticity in SAC, thus we experiment treating the temperature as both fixed and learnable parameter. We select 0.2 as the fixed temperature value for SAC. We get the performance curve by seeding 10 experiment from 0-9.

\begin{figure}[hbt!]
\centering
    \includegraphics[width=0.48\linewidth]{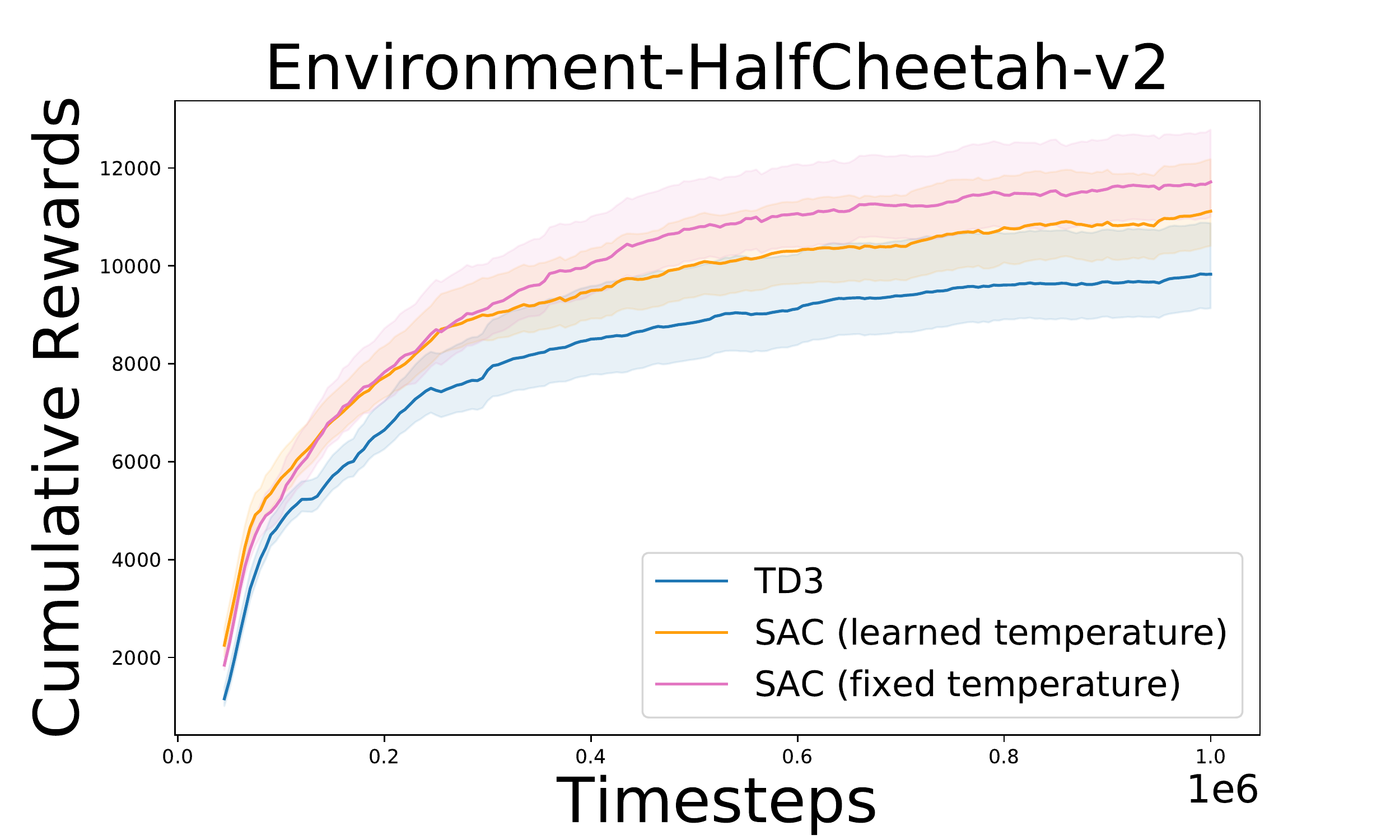}
   \!
    \includegraphics[width=0.48\linewidth]{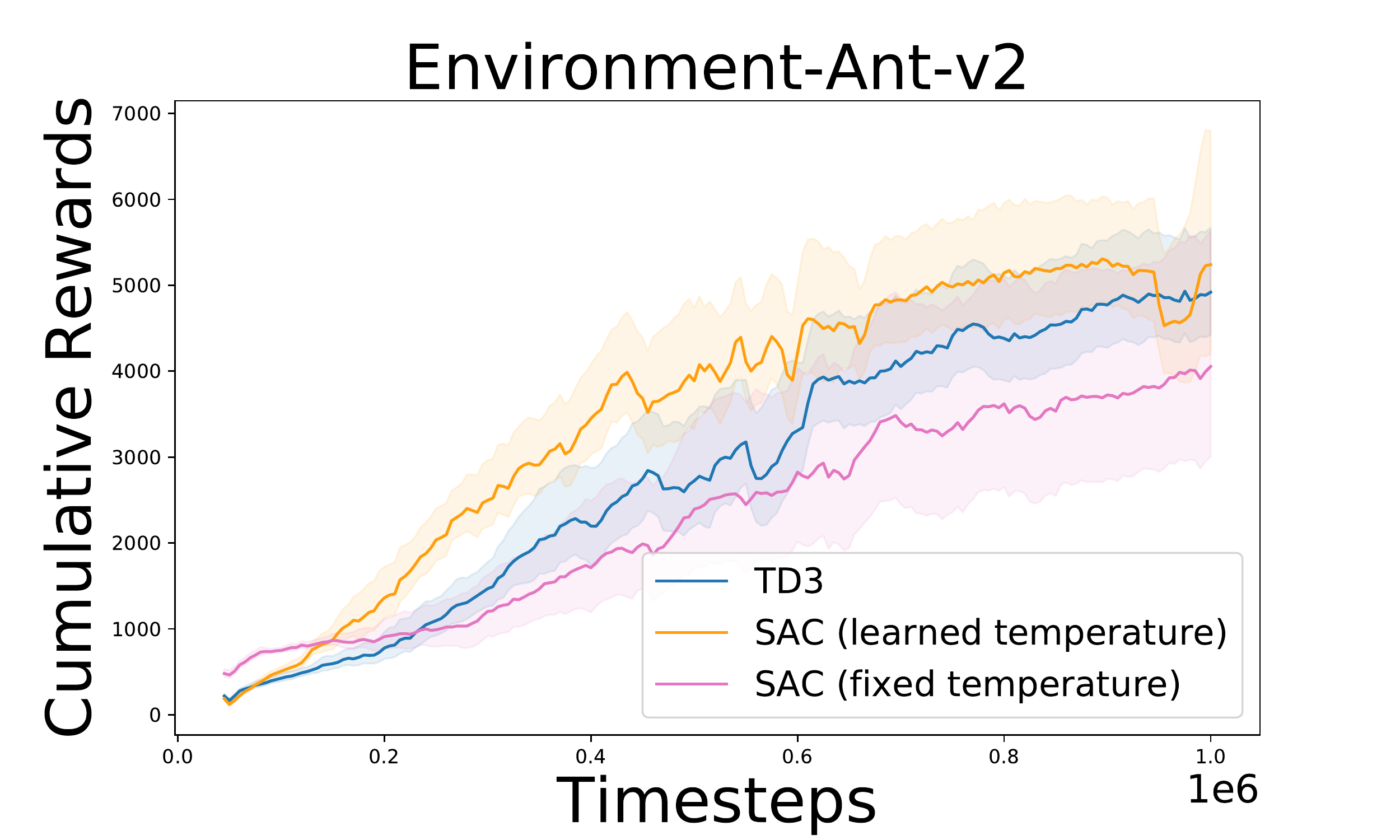}
   \!
    \includegraphics[width=0.48\linewidth]{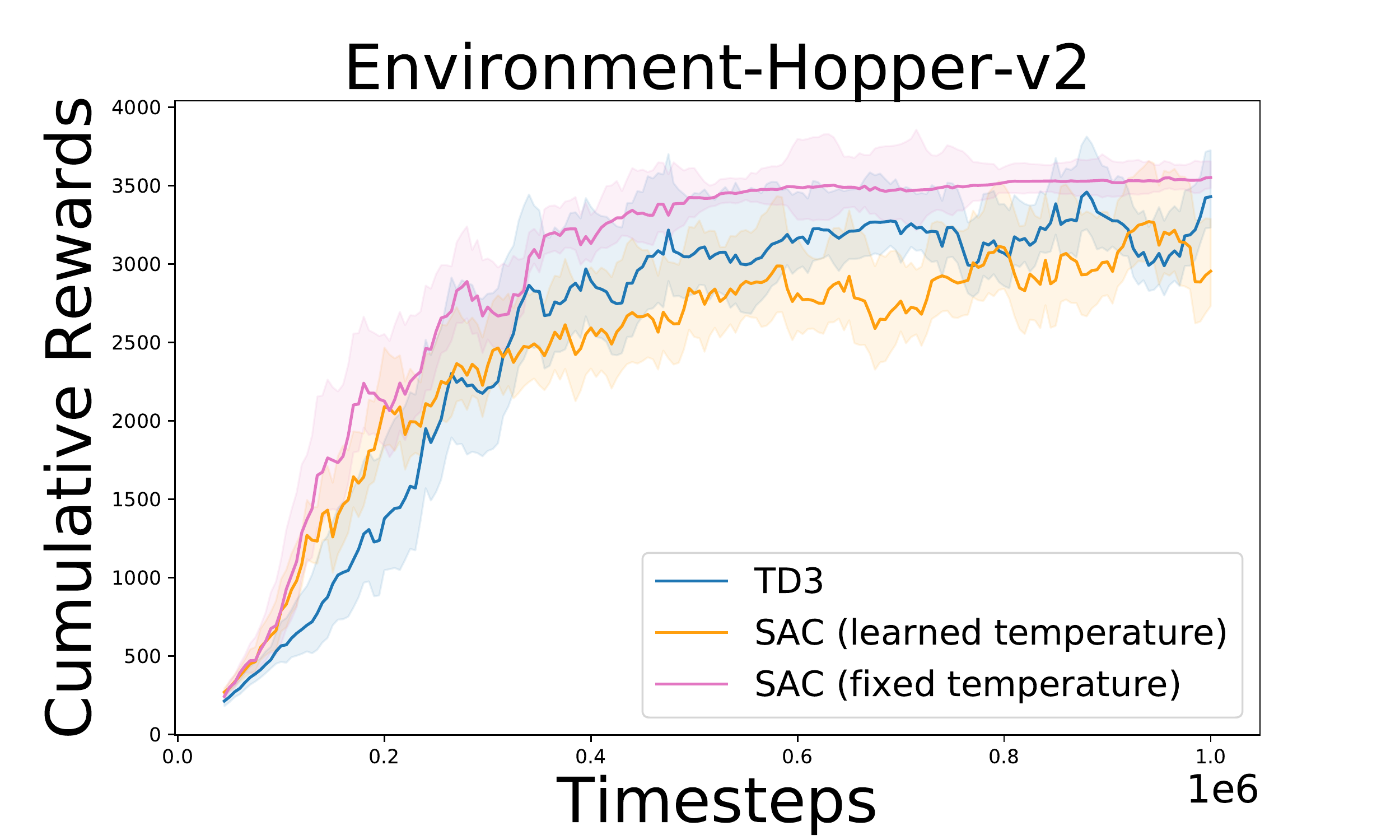}
   \!
    \includegraphics[width=0.48\linewidth]{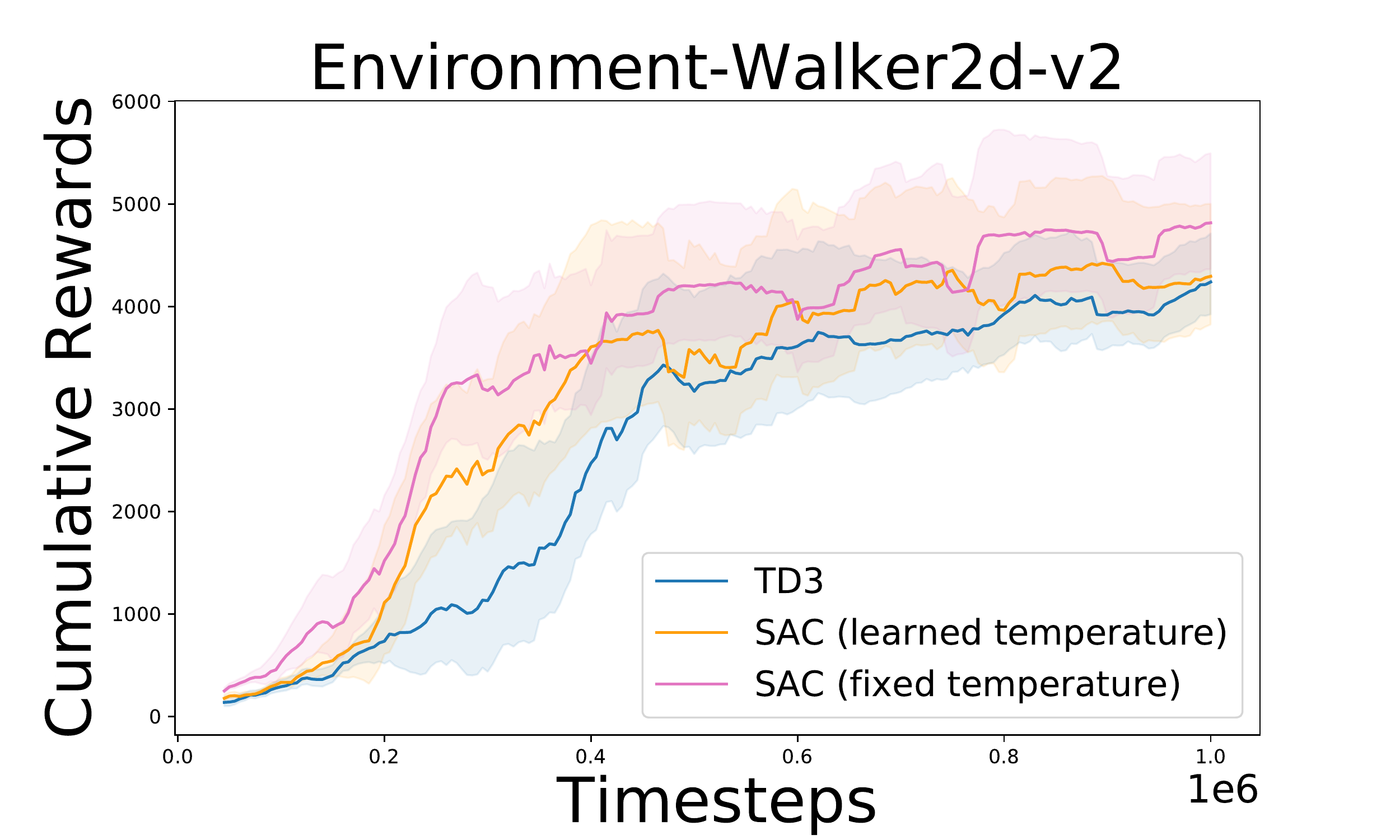}
    \!
    \includegraphics[width=0.48\linewidth]{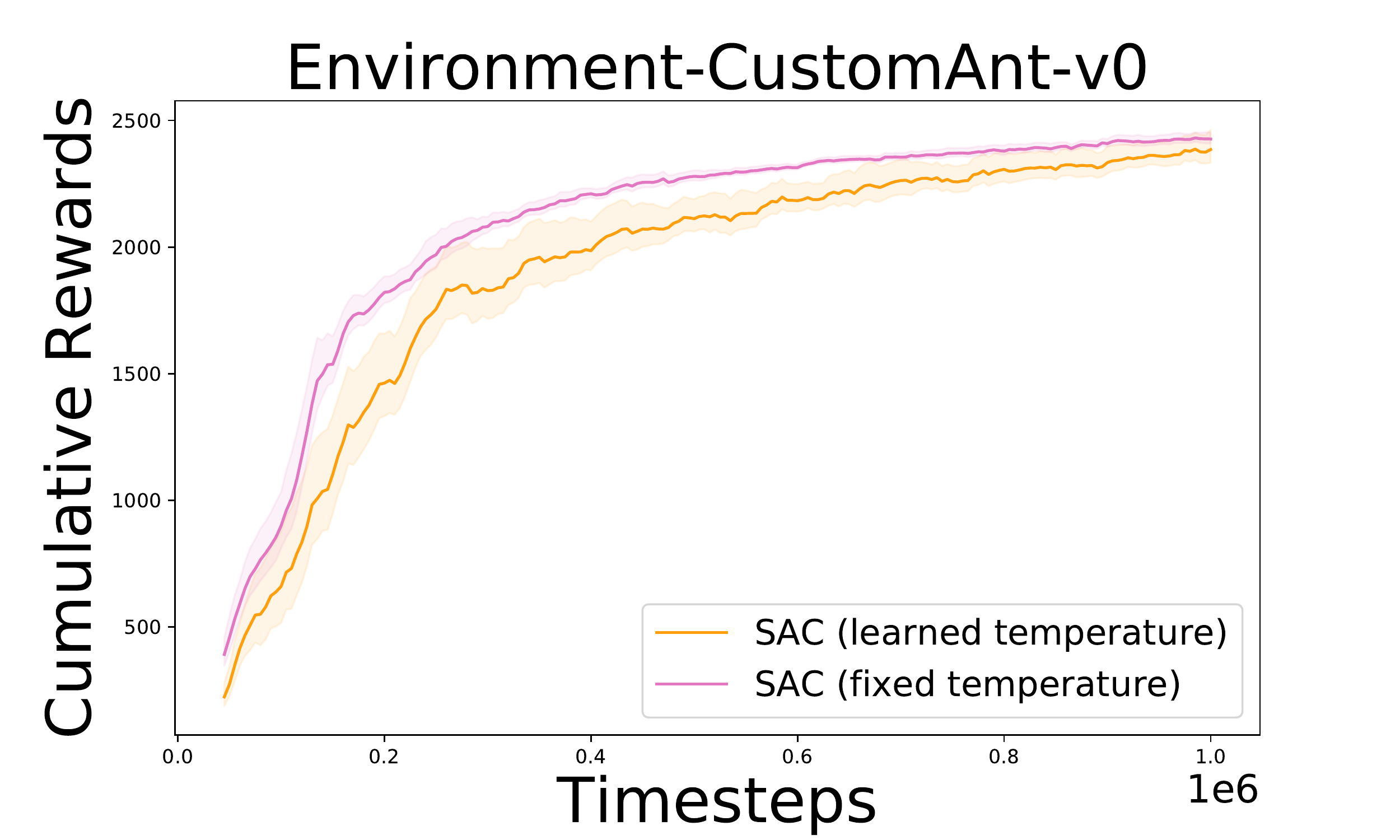}
   \!
    \includegraphics[width=0.48\linewidth]{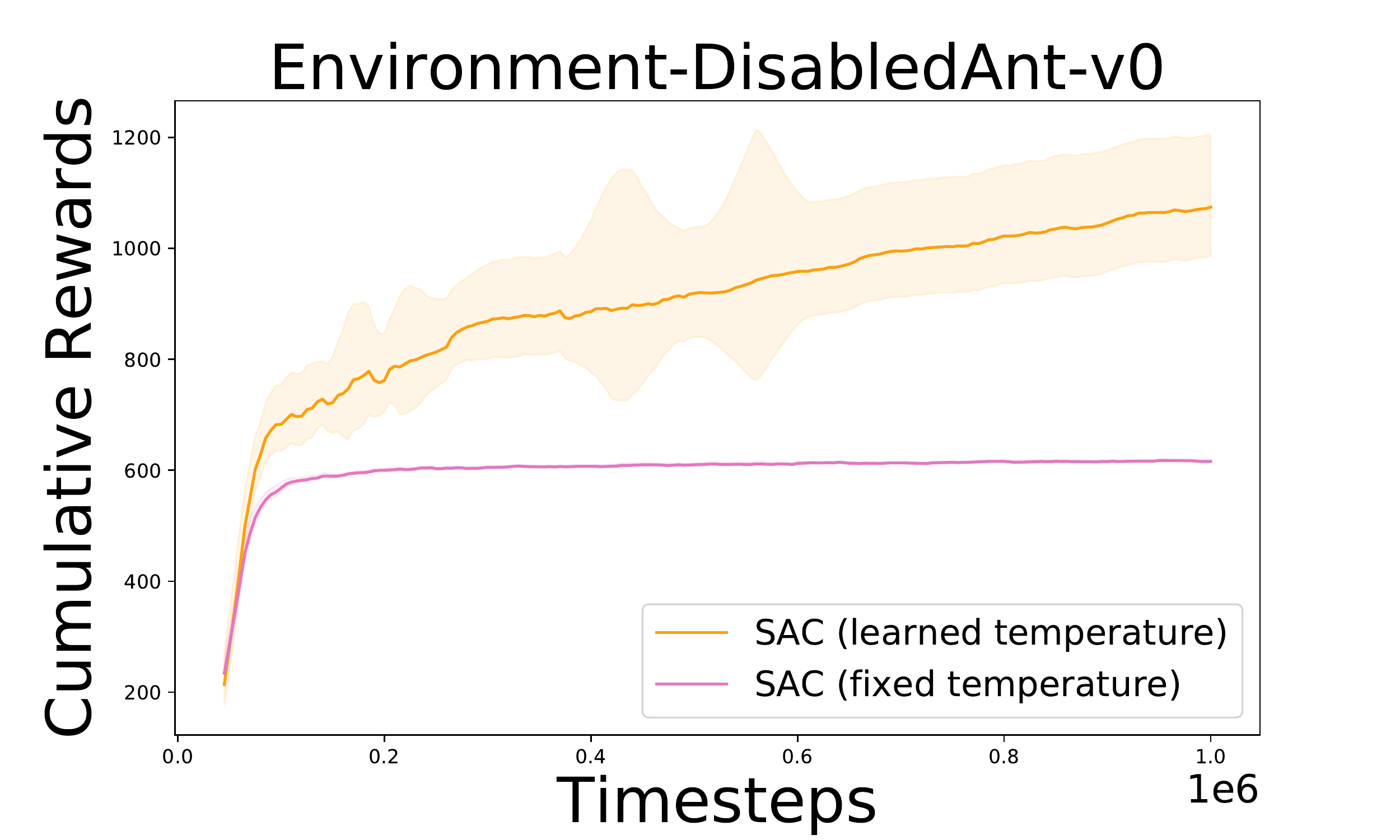}
   \!
    \includegraphics[width=0.48\linewidth]{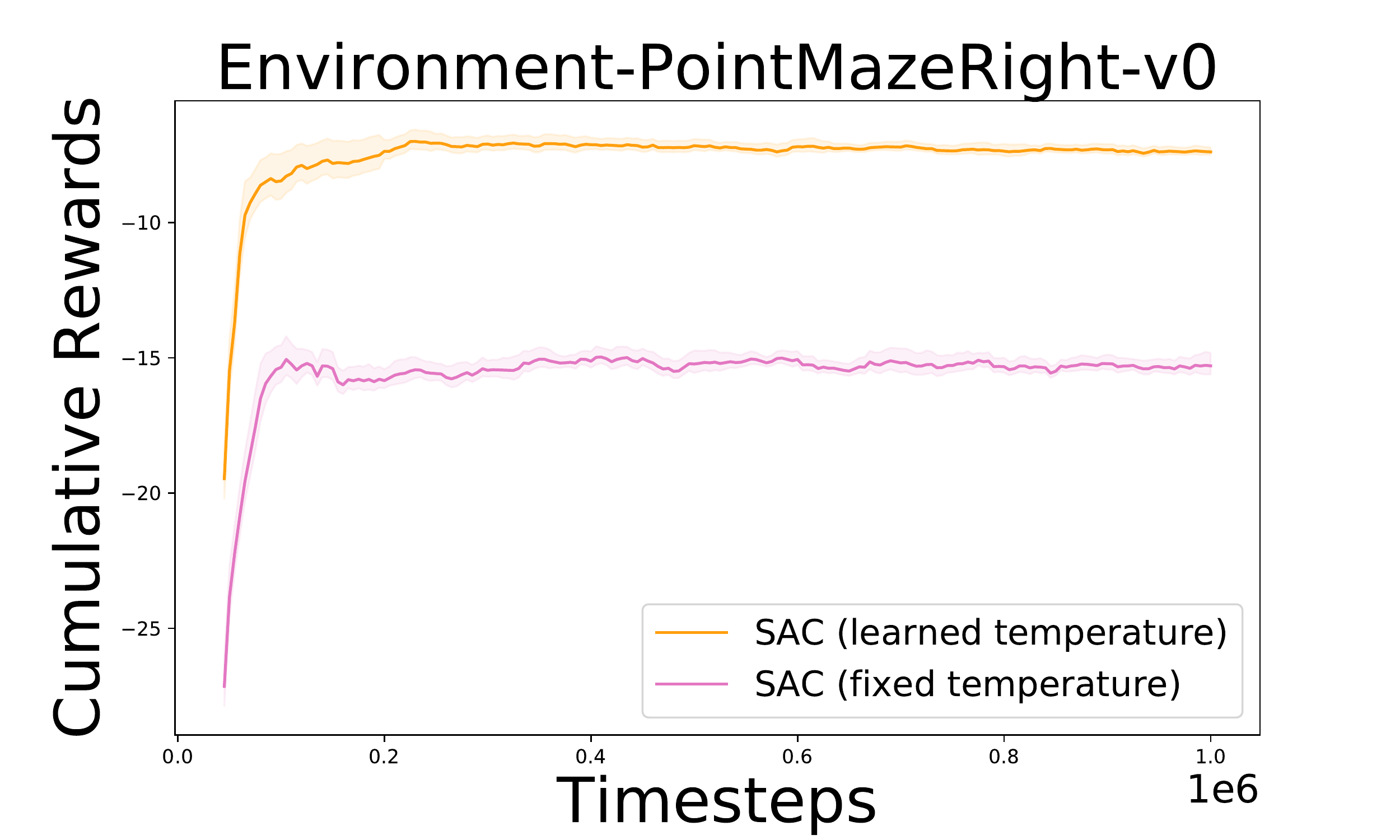}
   \!
    \includegraphics[width=0.48\linewidth]{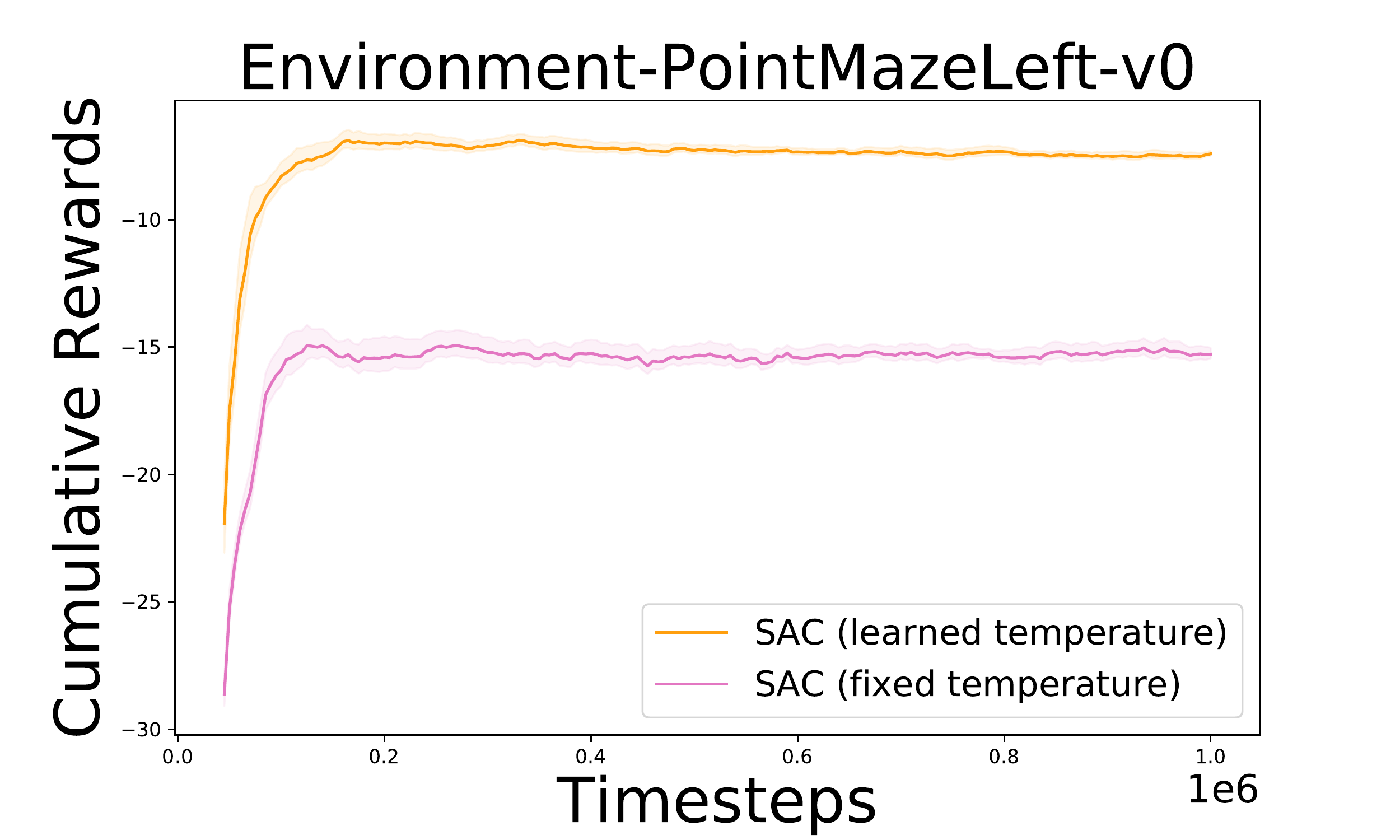}
    
%       \!
%     \includegraphics[width=0.4\linewidth]{SelectExpert/Swimmer-v2_RewardCurve.pdf}
%   \!
%     \includegraphics[width=0.4\linewidth]{SelectExpert/Humanoid-v2_RewardCurve.pdf}
\caption{Performance of policy gradient algorithms (TD3, SAC) over continuous control task in RL setting.}
\label{expert_selection}
\end{figure}

As can be seen from Figure \ref{expert_selection}, except for Ant-v2 in MuJoCo tasks, keeping the temperature value fixed for SAC gives better performance and for \cite{AIRLfu2018} environments SAC with learned temperature seem to perform with marginal improvement. Thus we select SAC as our expert for all the following experiments.

We collect 50 trajectories from 5 best performing expert policies. As the policy update is done by random sampling from the buffer, that itself creates randomness and help the learning policy to better generalize on the task \cite{DQN}. We use the same hyper-parameters (i.e. temperature) that gives the best results for the expert for all the experiments discussed in this paper.

% plotting details

\begin{table}[h!]
\centering
\begin{tabular}{|l|c|r|}
	\hline
     Environment &  Avg returns over seed 0-9 & Temperature parameter\\
	\hline
    HalfCheetah-v2 & 11083 & fixed 0.2 \\
    \hline
    Ant-v2 & 4294 & learned  \\
    \hline
    Hopper-v2 & 3544 & fixed 0.2 \\
    \hline 
    Walker2d-v2 & 4490 & fixed 0.2\\
	\hline
    CustomAnt-v0 &  2430 & fixed 0.2 \\
    \hline
    DisabledAnt-v0 & 1089 & learned  \\
    \hline
    PointMazeLeft-v0 & -7.37 & learned  \\
    \hline 
    PointMazeRight-v0 &  -7.33 & learned \\
	\hline
\end{tabular}
\caption{Expert Average Performance over 0-9 seed after $1e^6$ iterations.}
\label{expert_performance}
\end{table}

%%%%%%%%%%%%%%%%
%%%% figure %%%%
%%%%%%%%%%%%%%%%
\subsubsection{Performance of DAC with Different Policies}
% ==============================================

\begin{figure}[hbt!]
\centering
    \includegraphics[width=0.48\linewidth]{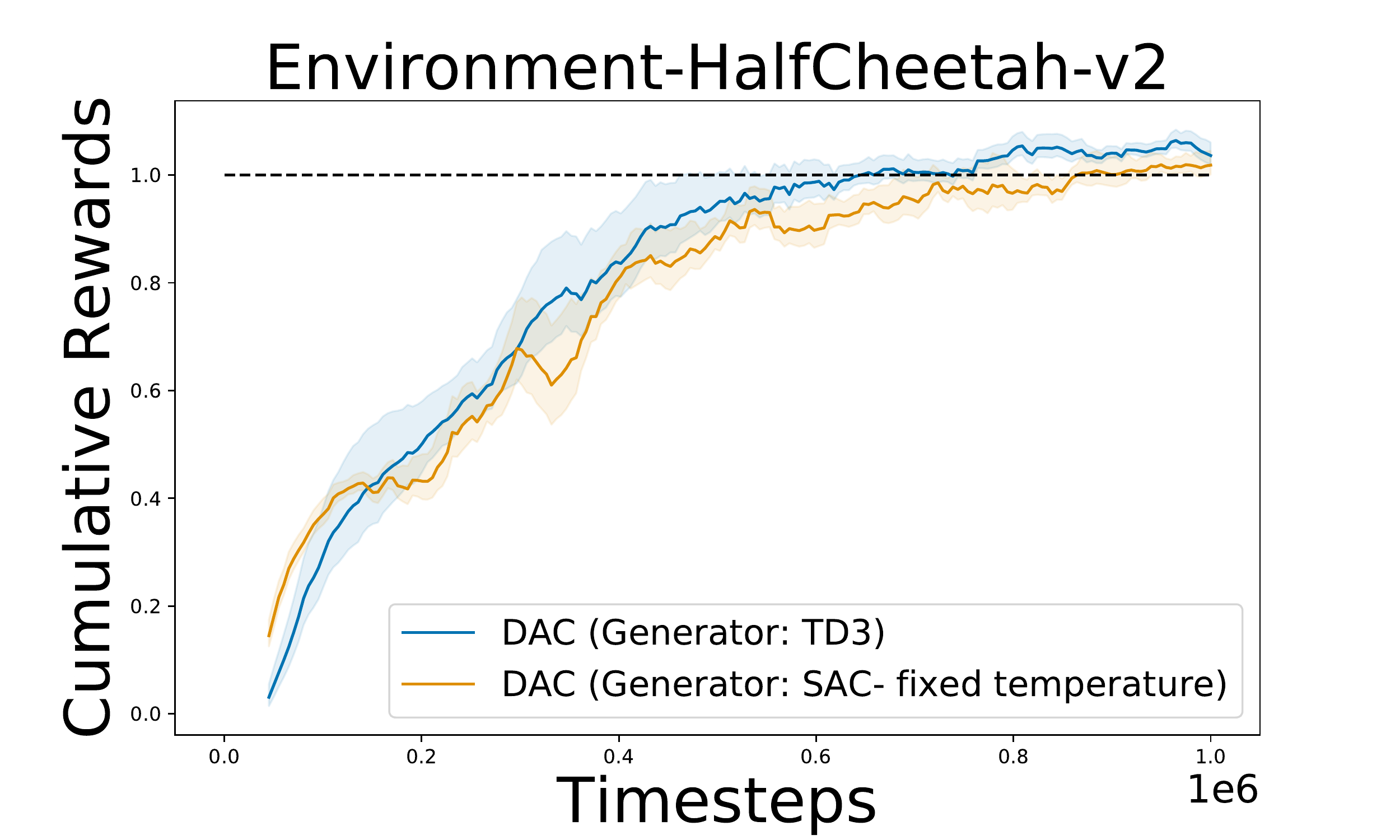}
   \!
    \includegraphics[width=0.48\linewidth]{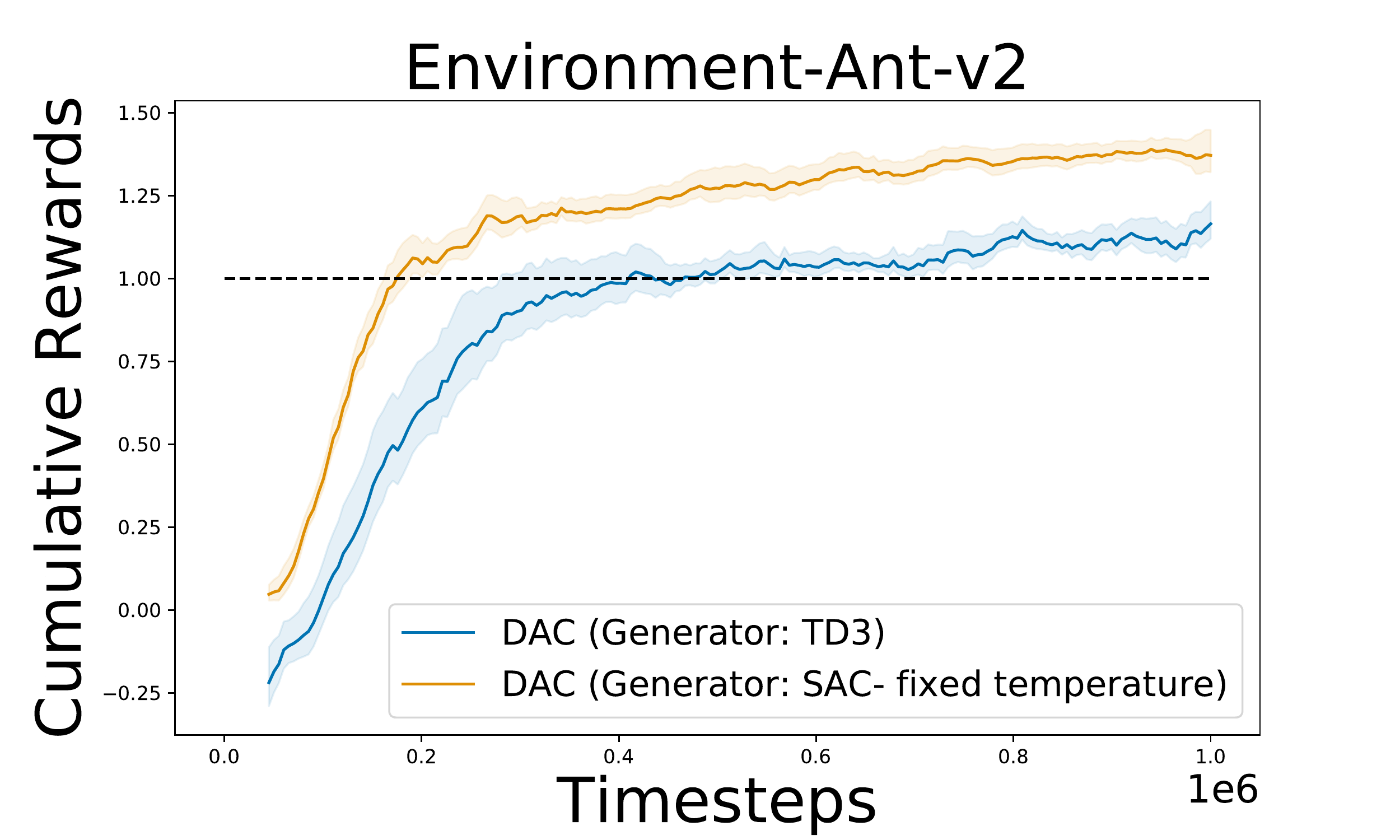}
   \!
    \includegraphics[width=0.48\linewidth]{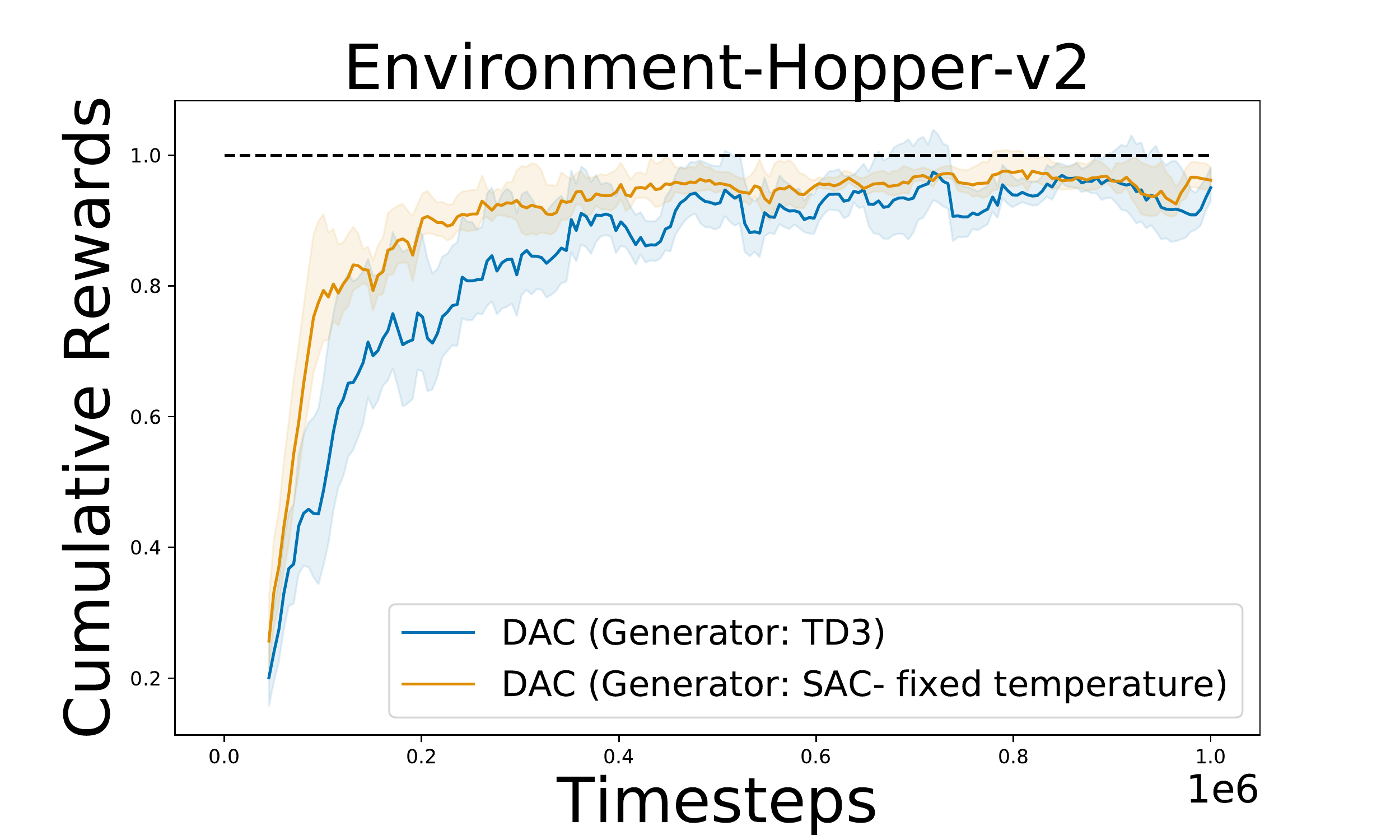}
   \!
    \includegraphics[width=0.48\linewidth]{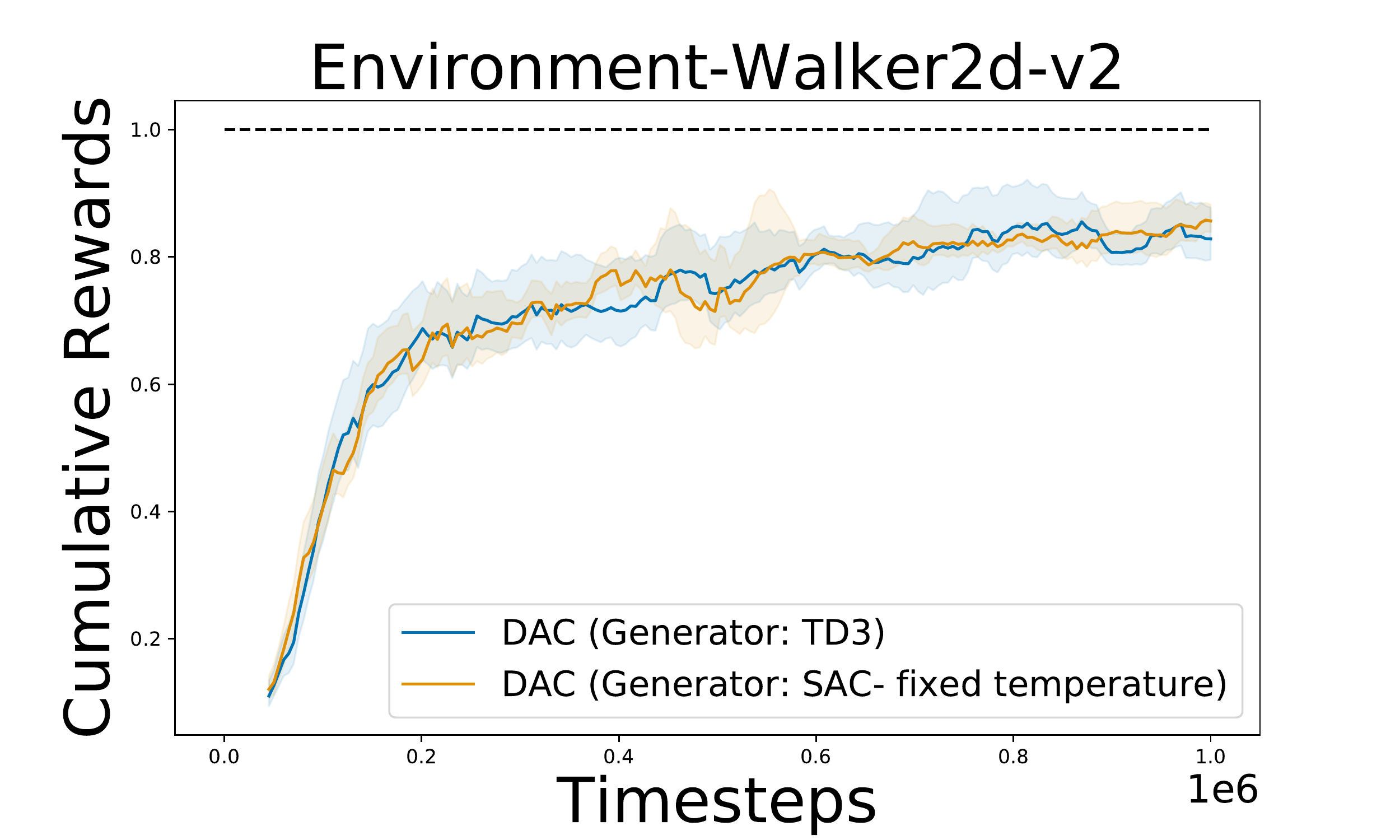}
\caption{Performance of DAC for different generator (TD3, SAC).}
\label{DAC_TD3vsSAC}
\end{figure}

No ablation study is conducted in \cite{DACkostrikov2018} with different policy as generator. Thus we also investigate the performance of DAC with \emph{SAC} as generator. SAC helps us with providing a entropy regularized objective function like GAIL \cite{GAILHoE16}. Our experiment using SAC in Figure \ref{DAC_TD3vsSAC} also proves the sample efficiency of DAC. Also it is important to have a stochastic policy to compute discriminator output ($D_{\psi,\omega} = \frac{ e^{f_{\psi,\omega}(s,a,s')} }{ e^{f_{\psi,\omega}(s,a,s')} + \pi_\theta(a|s)}$) in our off-policy-AIRL algorithm.  

% =====
% figure: DAC: TD3 vs SAC temp 0.2 
% =====

% ==========================================================
\subsection{Further details to compute premitives in MCP}
% ==========================================================

Each primitive $\pi_i(a|s,g) = \mathcal{N} \big( \mu_i(s,g),\Sigma_i(s,g) \big)$ is modeled by a Gaussian with mean $\mu_i(s,g)$ and diagonal covariance matrix $\Sigma_i(s,g) = diag\big( \sigma_i^1(s,g), \sigma_i^2(s,g) ... \sigma_i^{| \mathcal{A}|}(s,g) \big)$, where $\sigma_i^j(s,g)$  denotes variance of $j^{th}$ action parameter from primitive $i$ and $|\mathcal{A}|$ represents the dimensionality of the action space. A multiplicative comparison of Gaussian primitives yields yet another Gaussian policy $\pi(a|s,g) = \mathcal{N} \big( \mu_i(s,g),\Sigma_i(s,g) \big)$. Since the primitives construct each action parameter with an independent Gaussian, the action parameters of the composite policy $\pi$ will also assume the form of independent Gaussians with component-wise mean $\mu^j(s,g)$ and variance $\sigma^j(s,g)$. Component-wise mean and variance can be written as:

\begin{align}
    \mu^j(s,g) = & \frac{1}{\sum_{l=1}^k \frac{w_l(s,g)}{\sigma_l^j(s,g)} } \sum_{i=1}^k \frac{w_i(s,g)}{\sigma^j_i(s,g)} \mu_i^j(s,g), \\
     \sigma^j(s,g) = & (\sum \frac{w_i(s,g)}{\sigma^j_i(s,g)})^{-1}.
\end{align}

%===================

\end{document}